\documentclass[10pt,journal,compsoc]{IEEEtran}

\ifCLASSOPTIONcompsoc
  \usepackage[nocompress]{cite}
\else
  \usepackage{cite}
\fi
\ifCLASSINFOpdf
\else
\fi
\usepackage{xcolor}
\usepackage{helvet}
\usepackage{courier}
\usepackage{graphicx}
\usepackage{subfigure}
\usepackage{balance}
\usepackage{amsmath,amsthm,amssymb}
\usepackage{textcomp,booktabs}
\usepackage{footmisc}
\usepackage[normalem]{ulem}
\usepackage{multirow}
\usepackage{setspace}
\usepackage{bigstrut}
\usepackage{array}
\usepackage{xcolor}
\usepackage{amsfonts}
\usepackage{bm}
\usepackage{url}

\newcommand{\x}{\boldsymbol{x}}

\newcommand{\Dm}{\mathcal{D}}

\IEEEpubid{
\fbox{
\parbox{0.975\textwidth}{
\copyright 2021 IEEE. Personal use of this material is permitted. Permission from IEEE must be obtained for all other uses, in any current or future media, including\hfill
reprinting/republishing this material for advertising or promotional purposes, creating new collective works, for resale or redistribution to servers or lists, or reuse of\hfill
any copyrighted component of this work in other works. \vspace{-0.5mm}
}%
}}

\begin{document}

\title{Generalized Domain Conditioned \\Adaptation Network}
%
%
%
%

\author{Shuang Li, Binhui~Xie, Qiuxia~Lin, Chi Harold Liu,~\IEEEmembership{Senior Member,~IEEE}, Gao Huang and Guoren~Wang
\IEEEcompsocitemizethanks{
\IEEEcompsocthanksitem S. Li, B. Xie, Q. Lin, C. H. Liu and G. Wang are with the School of Computer Science and Technology, Beijing Institute of Technology, Beijing, China. Corresponding author: C. H. Liu. Email: \{shuangli, binhuixie, linqiuxia, chiliu, wanggrbit\}@bit.edu.cn\protect\\
\IEEEcompsocthanksitem G. Huang is with Department of Automation, Tsinghua University, Beijing, China, Email: gaohuang@tsinghua.edu.cn \protect\\
}
}

\markboth{IEEE transactions on pattern analysis and machine intelligence}
{Shell \MakeLowercase{\textit{et al.}}: Bare Advanced Demo of IEEEtran.cls for IEEE Computer Society Journals}
%



\IEEEtitleabstractindextext{%
\begin{abstract}
Domain Adaptation (DA) attempts to transfer knowledge learned in the labeled source domain to the unlabeled but related target domain without requiring large amounts of target supervision. Recent advances in DA mainly proceed by aligning the source and target distributions. Despite the significant success, the adaptation performance still degrades accordingly when the source and target domains encounter a large distribution discrepancy. We consider this limitation may attribute to the insufficient exploration of domain-specialized features because most studies merely concentrate on domain-general feature learning in task-specific layers and integrate totally-shared convolutional networks (convnets) to generate common features for both domains.
In this paper, we relax the completely-shared convnets assumption adopted by previous DA methods and propose \textit{Domain Conditioned Adaptation Network (DCAN)}, which introduces domain conditioned channel attention module with a multi-path structure to separately excite channel activation for each domain.
Such a partially-shared convnets module allows domain-specialized features in low-level to be explored appropriately. Further, given the knowledge transferability varying along with convolutional layers, we develop \textit{Generalized Domain Conditioned Adaptation Network (GDCAN)} to automatically determine whether domain channel activations should be separately modeled in each attention module. Afterward, the critical domain-specialized knowledge could be adaptively extracted according to the domain statistic gaps. As far as we know, this is the first work to explore the domain-wise convolutional channel activations separately for deep DA networks. Additionally, to effectively match high-level feature distributions across domains, we consider deploying feature adaptation blocks after task-specific layers, which can explicitly mitigate the domain discrepancy. Extensive experiments on four cross-domain benchmarks, including DomainNet, Office-Home, Office-31, and ImageCLEF, demonstrate the proposed approaches outperform the existing methods by a large margin, especially on the large-scale challenging dataset.
The code and models are available at \url{https://github.com/BIT-DA/GDCAN}.

\end{abstract}
}

\maketitle

\IEEEdisplaynontitleabstractindextext

%
\IEEEpeerreviewmaketitle

\ifCLASSOPTIONcompsoc
\IEEEraisesectionheading{\section{Introduction}\label{sec:introduction}}
\else
\section{Introduction}
\label{sec:introduction}
\fi

\IEEEPARstart{C}{onvolutional} Neural Networks (CNNs) have played a predominant role in various visual applications~\cite{alexnet,resnet,Long2014Fully,Simon2014,densenet_pami} by seeking hierarchical representations. However, there are two essential pre-requisites for effective performance: large-scale labeled training data~\cite{imagenet-dataset} and the same/similar distribution across training and test datasets~\cite{CDAN,JADA}. Unfortunately, in real-world scenarios, obtaining sufficient labeled data through manually labeling is time-consuming or downright infeasible. Also, it is impractical to expect that test and training data share an identical distribution. The reason for this dilemma is that the data often comes from different domains, such as training images might be carefully selected without complex background while test images could be camera snapshots taken anywhere. If the difference across domains can be eliminated, the network trained by labeled source could be generalized well to unlabeled target.

As an effective strategy to implement this idea, domain adaptation (DA) is gathering momentum in the past decade \cite{survey,DICD,DICE}. Early DA methods generally aim to align domain distributions by either reweighting instances~\cite{KMM,PRDA} or learning domain-invariant features~\cite{DTLC,SCA,JDA}. Subsequently, given the powerful feature extraction ability of CNNs, numerous deep DA works have been discussed to boost performances in a variety of tasks~\cite{blitzer2006domain,zhu2017unpaired,Zhang2017seg,DeCAF}. Among them, cross-domain image classification is a classical and representative problem in computer vision, and there are basically two kinds of strategies for this practical DA problem: domain discrepancy minimization~\cite{JAN,MDD,DAN_pami,DRCN,DCORAL,DDC} and adversarial learning~\cite{ADDA,MADA,CDAN,MCD,SPCAN,DANN}. Their goals are to reduce domain shift in the top task-specific layers to make features more transferable. However, they usually assume the convolutional layers are universal across different domains in capturing general low-level features based on the analysis in~\cite{transferable}.

As a matter of fact, this basic assumption brings out two limitations. Firstly, the aforementioned methods are to find those representations using an \textit{identical convolutional architecture} for both domains. Intuitively, however, as the domains have different properties, the design that all channels in convnets are of equal importance is unfit for DA, especially when the domain discrepancy is tremendous. Here, we take the task of pnt $\rightarrow$ skt in DomainNet as an example. Painting (pnt) and Sketch (skt) are two domains of DomainNet, by far the largest and most challenging domain adaptation dataset. Painting contains artistic depictions of objects in the form of paintings while Sketch contains sketches of specific objects. When using an identical convolutional architecture for both domains, the convolutional filters would be more exclusively sensitive to source specialized features (i.e., color and style) because of the source supervised training, and fail to capture domain-informative features for target data (i.e., outline and shape). It is clear that some channels are easier to transfer than others since they constitute the sharing patterns of both domains.
Therefore, a natural approach is to allow domains to undergo partially-different architectures to  preserve domain-specific information while arriving at domain-invariant feature representations.

Secondly, only deploying domain discrepancy penalty terms or adversarial losses at top layers may be less effective, since the gradients of the loss modifications at the task-specific layers will be decayed via the back-propagation scheme.
As a result, the shared convolutional layers across domains may lose domain-specialized knowledge at the beginning of the very deep convolutional networks.

\begin{figure}[tb]
\centering
\includegraphics[width=0.99\columnwidth]{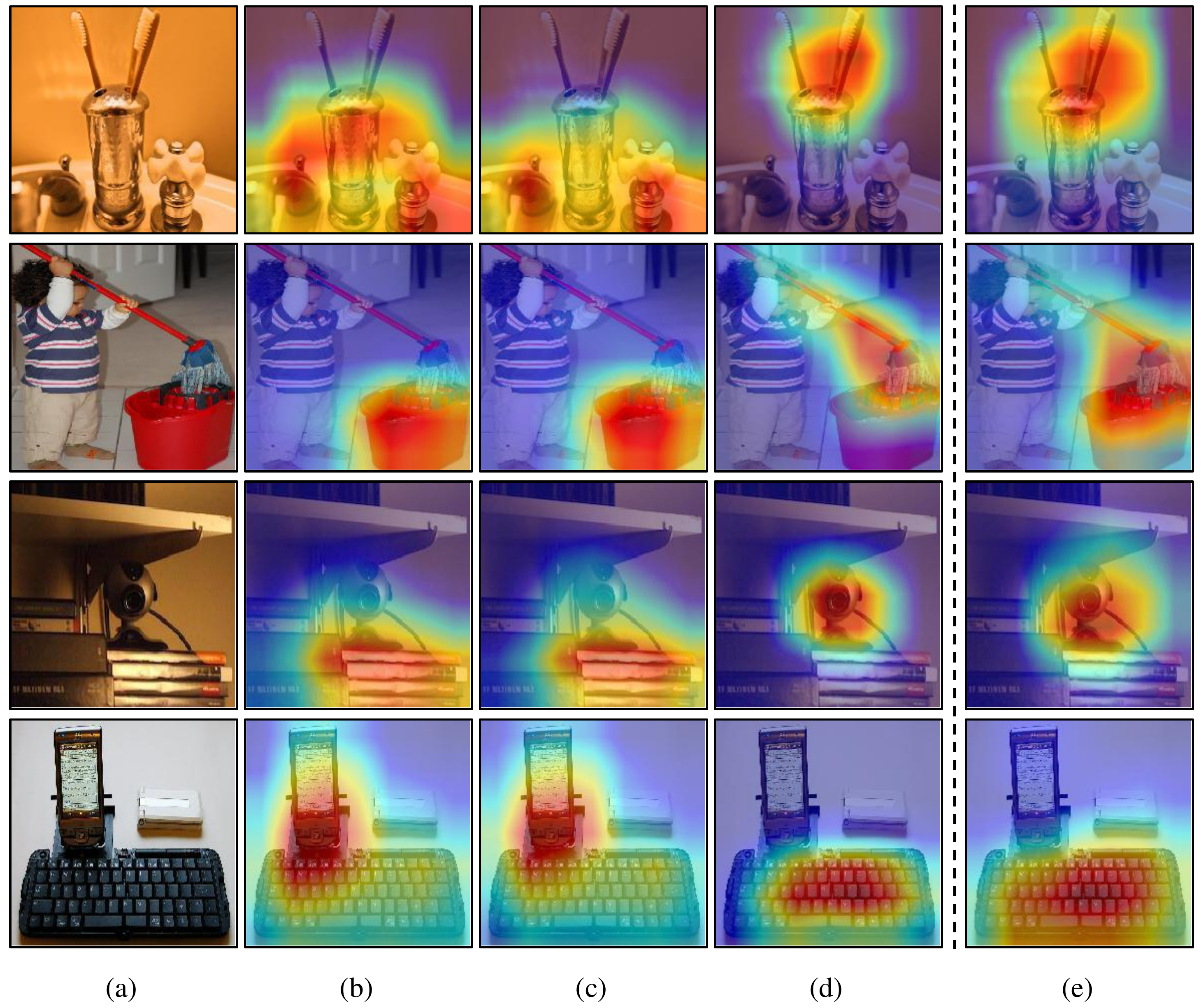}\vspace{-4mm}
\caption{(Best viewed in color.) Attention visualization of the last convolutional layer of different models on the task Ar $\rightarrow$ Rw of Office-Home. The first column (a) shows the original target images from randomly selected classes (i.e., {toothbrush}, {mop}, {webcam}, and {keyboard}); other columns show the attention maps from (b) source-only model, (c) DCAN w/o domain conditioned attention module, (d) DCAN, and (f) model with target ground-truth labels, respectively.
}\vspace{-4mm}
\label{Fig_attention_visualization}
\end{figure}

It is believed that the versatile feature representations should be able to reduce the difference of cross-domain distributions as much as possible, and simultaneously preserve specialized properties within domains.
To intuitively clarify our argument, we utilize Grad-CAM visualization \cite{grad-cam} to generate attention maps of different models and demonstrate that totally-shared convnets with distribution alignment loss may still mislead the predictions on target data. As shown in Fig. \ref{Fig_attention_visualization}, the source-only model in column (b) incorrectly localizes objects (i.e., bottle, bucket, notebook, and telephone) merely under the guidance of source supervision, which verifies the limitation of traditional deep learning methods without adaptation. However, in column (c), similar to most DA methods, only conducting distribution alignment
on the top task-specific layers causes misclassification as well. A reasonable interpretation is that their shared parameters of convents hinder domain-specific characteristic learning, especially when dealing with a large domain discrepancy.
Note that column (e) visualizes attention maps of the model trained with target ground-truth label, wherefore it shows the areas of greatest interest to the target domain. It is clear that the desirable regions (i.e., toothbrush, mop, webcam, and keyboard) can be consistently highlighted by DCAN in (d), a partially-shared convnets framework, whose results are similar to that of column (e). Consequently, we conclude that it is crucial to explore an effective domain-specialized convolutional representation learning mechanism to seize the core information within domain, which is a key ingredient of our approach.

Based on the aforementioned discussions, in this paper, we develop \textit{Domain Conditioned Adaptation Network (DCAN)} to address unsupervised visual domain adaptation problem, which sheds new light on domain-specialized feature learning in convolutional layers.
Specifically, we design a lightweight domain-conditioned channel attention module in convnets, which averages channel statistics to gather global information of each domain and feeds them into their respective activation paths according to the domain label. With domain-dependent knowledge being fully exploited, we allow large deviations exist in feature representations of different domains. This could improve the representation power and flexibility of the network.

Meanwhile, regarding bottom layers in deep CNNs usually encode low-level general features that are lack of domain discrimination~\cite{transferable}, it is unreasonable to enforce route separation for each domain in all attention modules. Ideally, if the statistic differences across domain representations are small, making source and target domains share identical channel activation structure could better improve the transferability of deep features as stated in~\cite{transferable}. Therefore, we further propose \textit{Generalized Domain Conditioned Adaptation Network (GDCAN)} to adaptively model domain channel individual activations in each attention module according to their statistic differences. Different from most DA methods only deploying discrepancy loss at top layers, we additionally plug feature adaptation modules into task-specific layers with a simple regularizer, which can explicitly reduce the domain shift. As a consequence, our method offers a promise that domain-specialized features in low levels would be preserved while domain-general features can be sufficiently learned in higher levels.
To summarize, our contributions can be concluded as follows:
\begin{itemize}
    \item Firstly, we propose a partially-shared structure in convnets leveraging domain conditioned channel attention module, which divides processing data stream into source and target routes. Such a multi-path scheme allows each domain to perform feature recalibration separately and the representations at low-level layers are expected to be domain-informative. Further, we extend it to an adaptive routing strategy to independently model domain-specialized channel activation in each convnet block.
    \item Secondly, we incorporate feature adaptation modules in all task-specific layers to learn domain-invariant representations more effectively. To assist the target domain in better adapting to the source domain, we also introduce an additional regularizer to properly guide the learning of the feature adaptation module.
    \item Thirdly, as general methods, the channel attention and feature adaptation module in DCAN and GDCAN can be easily applied to other popular deep architectures and domain adaptation methods to further improve their transferability.
    \item Finally, we conduct comprehensive experiments on four standard benchmarks, including DomainNet, Office-Home, Office-31, and ImageCLEF-DA. Our method outperforms all comparisons with significant improvements. Particularly, for the most challenging dataset, DomainNet, the average classification accuracies of our methods outperform the best baseline over $8\%$, bringing the performance to a new level.
\end{itemize}

A preliminary version of this work was presented in the conference paper \cite{DCAN}. In this extension we mainly make the following improvements:
(1) We build upon conference version DCAN and put forward GDCAN via applying an adaptive channel attention module within convolutional networks depend on the cross-domain statistic differences;
(2) For practical DA challenges, we provide clear insight and necessity of selectively sharing channel attention activation branch in convolutional layers;
(3) Given the universality of our approaches, we integrate the designed modules of DCAN and GDCAN into other deep architectures and comparable DA approaches to further boost their adaptation capabilities;
(4) We further enlarge the experimental parts by evaluating DCAN and GDCAN on more public benchmark datasets including all-task DomainNet, and design comprehensive analysis to carefully verify the superiority of the adaptive version GDCAN.

\section{Related Work}\label{sec:relatedwork}
This section reviews related deep DA works, mainly covering: discrepancy-based, adversarial-based, attention-based, and domain-specialized architecture-based methods.
\vspace{-2mm}
\subsection{Discrepancy Metric Minimization}
Classical domain adaptation methods devote to seeking domain-invariant features in task-specific layers through various statistical moment matching techniques~\cite{DCORAL,CMD,MMD}. Among them, Maximum Mean Discrepancy (MMD)~\cite{MMD} is one commonly-used criterion. Long et al. explore multi-kernel MMD (MK-MMD) to minimize marginal distributions of two domains in~\cite{DAN}. JAN in~\cite{JAN} introduces JMMD that aligns joint distributions of multiple layers.
Another example is RTN~\cite{RTN} that considers feature fusion with MMD and designs a residual function to perform classifier adaptation. Further, DRCN in ~\cite{DRCN} utilizes residual correction block to explicitly mitigate the domain feature gap. Apart from MMD, Zhang et al.~\cite{MDD} define a new divergence, Margin Disparity Discrepancy (MDD), and validate that it has rigorous generalization bounds. SAFN in \cite{SAFN} not only utilizes norm to quantitatively measure domain statistics but also suggests that larger norm features can boost knowledge transfer.

Unfortunately, all these works enforce source and target data to share one identical backbone convolutional network, which usually underestimates the domain mismatch in the low-level convolutional stage.
Moreover, as the learning ability of task-specific layers cannot compensate the side effect of over-shifting extraction in low-level convolutional layers, the performance of DA methods would be limited when the source and target domains differ to a large extend.

\vspace{-2mm}
\subsection{Adversarial Learning}
Another route of research is exploring domain-invariant representations by a two-player minimax game inspired by Generative Adversarial Networks (GANs) \cite{GAN}.
This class of methods uses a discriminator to distinguish source features from target features and learns a feature extractor capable of confusing the discriminator.
For example, DANN~\cite{DA_bp} attempts to learn task-specific domain-invariant features through a novel gradient reversal layer (GRL).
GTA~\cite{GTA} transfers target distribution information to the learned embeddings utilizing a generator-discriminator pair. In addition, MCD in~\cite{MCD} introduces a new adversarial paradigm by maximizing two classifiers' decision disagreement while training a generator to minimize it. On top of MCD DANN, Li et al.~\cite{JADA} leverage joint alignment to achieve more accurate results. Domain-symmetric networks (SymNets) in \cite{SymNets} apply an additional classifier to facilitate the alignment of joint distributions of feature and category via two-level domain confusion losses.
To capture multi-mode structures, CDAN \cite{CDAN} enables discriminative adversarial adaptation by conditioning target predictions.
Zhang et al. \cite{iCAN} design a collaborative network by adding several domain classifiers on multiple stages and extend it to SPCAN \cite{SPCAN} trained with weighted pseudo-labeled target samples. BSP in \cite{BSP} aims to alleviate the deterioration of feature discriminability in adversarial learning, presenting Batch Spectral Penalization. Cui et al. \cite{BNM} propose Batch Nuclear-norm Maximization (BNM) to jointly improve discriminability and diversity.

Despite their efficacy in various tasks, existing adversarial DA methods are hard to achieve stable solutions compared with discrepancy-based methods. This can be explained by their intrinsic adversarial training strategy, which is less effective when one side is much stronger.

\vspace{-2mm}
\subsection{Attention-based Methods}
Attention mechanism~\cite{Attention} enables a neural network to accurately focus on all the relevant elements of the input. There are mainly two attention mechanisms widely used in computer vision studies, spatial attention and channel attention, which aim to capture the pixel-level pairwise relationship and channel dependency, respectively.
To name a few, Squeeze-and-excitation network (SENet)~\cite{SE} adaptively recalibrates channel-wise feature representations by modeling channel correlations. Convolution block attention module (CBAM)~\cite{CBAM} improves SE block by additionally exploring spatial attention. Later on, Lee et al. propose a style-based recalibration module (SRM)~\cite{SRM} to extract style information for intermediate convolutional feature maps through style pooling.
Indeed, these techniques for supervised learning have received considerable attention. However, they have not been explored in domain adaptation due to the domain shift dilemma where the images across domains are drawn from different distributions. Different from them, we seek to determine the suitable attention activations for each domain separately to promote the model adaptation performance.

Besides, in a manner of attention alignment, it has been shifted to DA. For instance, Zhuo et al. \cite{DUCDA} propose an attention transfer process for convolutional domain adaptation with aligning attention maps for two domains.
Kang et al. \cite{DAAA} propose the deep adversarial attention alignment (DAAA) to transfer knowledge in all convolutional layers by attention matching. Considering local and global attentions, TADA in \cite{TADA} aims to highlight transferable regions or images across domains.

Different from them, our work turns to a channel attention mechanism which is a more simple and effective structure module. Besides, ours and the aforementioned methods serve two different purposes. Namely, they apply an attention map in order to detect transferable features for DA, while we leverage domain conditioned channel attention in convolutional layers with partially-shared parameters to activate distinctly interested channels for each domain. Therefore, our channel attention mechanism would benefit the representation learning of inter-domain features as well as domain-specific ones, making it more flexible and powerful in modeling complex data from different domains.

\vspace{-2mm}
\subsection{Domain-specialized Architecture Based Methods}
As the performance of DA methods is tightly linked to the network architectures, some works begin to design domain-specialized architectures that process the source and target data separately. Chang et al. present domain-specific batch normalization (DSBN)~\cite{DSBN} to learn domain-specific information for each domain separately. Later on, Carlucci et al.~\cite{AutoDIAL} introduce novel domain alignment layers (DA-layers), which automatically learn the degree of good alignment at different levels of the network. Li et al.~\cite{AdaBN} show that modulating the statistics from source domain to target domain in all batch normalization layers could achieve effective adaptability. Similarly, Wang et al.~\cite{transnorm} propose transferable normalization (TransNorm) module to replace the shared batch normalization, which enables CNNs more transferable. Roy et al.~\cite{DWT-MEC} design domain-specific whitening transform (DWT) layers after the convolutional layers for the purpose of matching two domains.

Most of the above methods aim to perform domain-specific standardization to the feature activations in normalization layers, while the correlation between activations within domains in convolutional layers has not been fully explored, leading to suboptimal adaptation efficacy. By contrast, our methods enforce source and target domains to extract domain-informative knowledge in low-level convolutional layers and indistinguishable representations in high-level task-specific layers by the designed domain conditioned attention and feature adaptation modules, which are more effective than only replacing the feature normalization modules. Moreover, these domain-specialized normalization techniques are orthogonal to the contribution of this work, which mainly focuses on learning domain conditioned activations in convolutional layers. In Section~\ref{sec:increment_exp}, we show that the proposed domain-specialized channel activation mechanism can be effectively integrated with other domain adaptation methods to further promote their generalization performance on the target domain.

\begin{figure*}[tb]
  \center
  \includegraphics[width=0.85\textwidth]{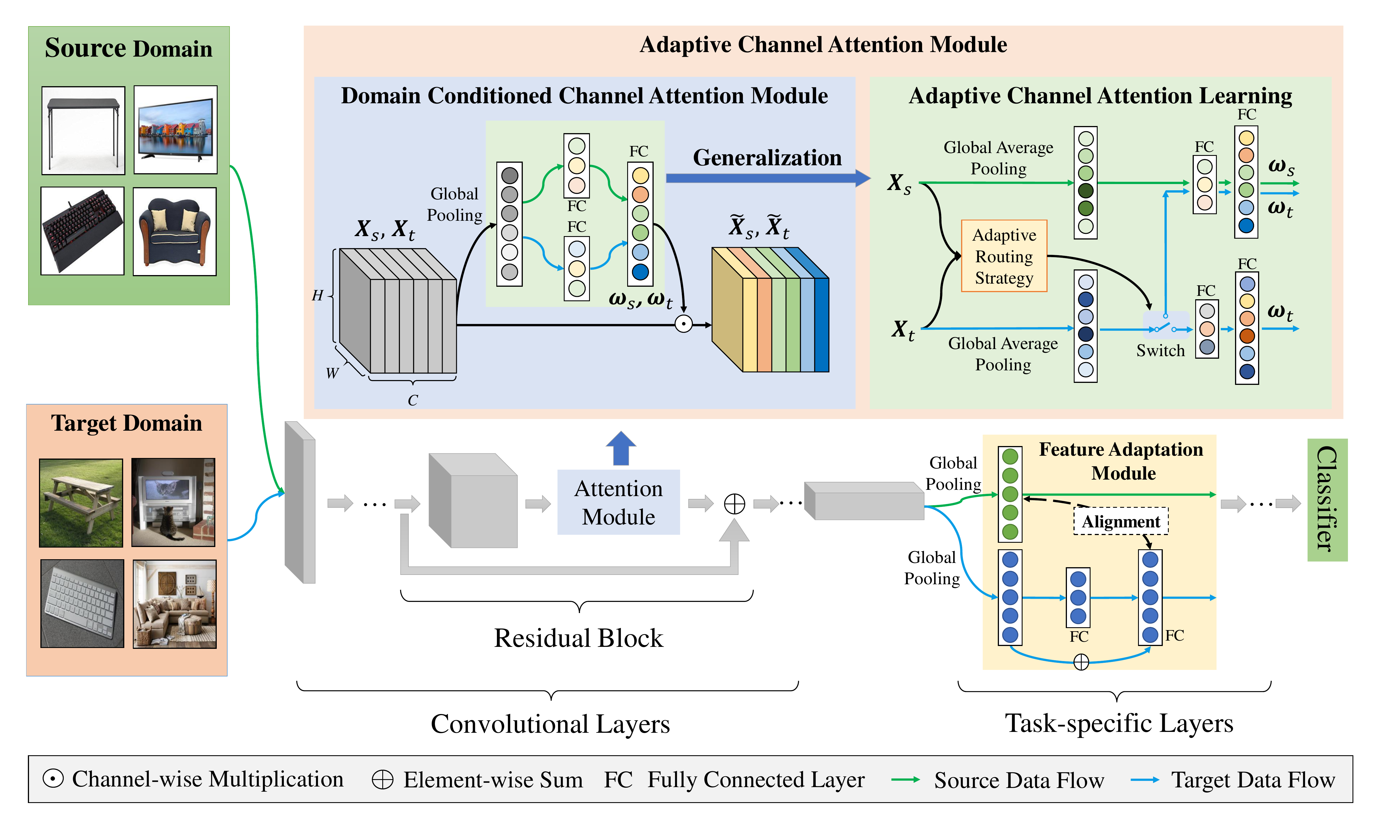}\vspace{-4mm}
  \caption{Overview of our proposed method. We introduce two effective modules into the network to transfer domain-specialized and domain-invariant features simultaneously. For the domain conditioned channel attention module, we add it into each residual block. It is a multi-path structure segregating source and target into different processing flows, which models fine-grained details of each domain.
  If applied with the adaptive routing strategy, this structure can be extended to the adaptive channel attention module, which determines the routing of target channel attention calculation based on the cross-domain statistic distance (shown in the \textit{right green part}).
  As for the high-level feature adaptation module, it is plugged into multiple task-specific layers and only target data are allowed to pass through it during the alignment. By aligning transformed target data and unchanged source data, we can make feature adaptation module explicitly measure domain discrepancy to derive more domain-invariant features.
   }
  \label{Fig_method}\vspace{-4mm}
  \end{figure*}

  \section{Method}\label{sec:method}
  \subsection{Notation and Preliminaries}
  
  To formalize the problem of unsupervised DA, we denote $\Dm_s=\{(\boldsymbol{x}_{1}^s, y_{1}^s), ..., (\boldsymbol{x}_{n_s}^s, y_{n_s}^s)\}$ as source domain and $\Dm_t=\{\boldsymbol{x}_{1}^t, ..., \boldsymbol{x}_{n_t}^t\}$ as target domain. In source domain $\Dm_s$, there are $n_s$ labeled samples, and a source pair is source sample $\boldsymbol{x}_{i}^s$ with its corresponding label $y_{i}^s$.
  As for target domain $\Dm_t$, a total of $n_t$ target samples are unlabeled.
  We assume they have the same label space, with $\mathcal{C}_n$ common classes.
  Since the distributions across two domains are different, i.e., $P_s \neq Q_t$, it is impossible to expect satisfactory performance on target tasks by directly imposing classifier trained on source data.
  Our goal is to generalize well on the target domain by exploring labeled source data and unlabeled target data in the training stage.
  
  Generally, discrepancy metric minimization and adversarial learning based methods seek domain-invariant feature representations by distance minimization or domain confusion.
  Influenced by the intuition of convolutional layers capturing common low-level features across various domains, they directly deploy completely-shared convnets for source and target. However, due to the interference of source supervised learning, we believe the shared convnets would trigger the degradation in target performance as it lacks specialized feature learning of the target domain.
  To make matters worse, we often encounter that source and target domains have a huge distribution discrepancy.
  Therefore, there is a strong motivation to design a weakly-shared structure so as not to lose the general learning ability of convnets while strengthening domain-wise feature learning.
  Except that, the cross-domain high-level features should also be explored to facilitate discriminative knowledge transfer.
  
  To this end, we propose a Domain Conditioned Adaptation Network (DCAN) to simultaneously capture domain-specific and general representations. As shown in Fig. \ref{Fig_method}, our framework consists of two main components: domain conditioned channel attention module in convolutional layers and feature adaptation module in task-specific layers.
  With the introduced channel attention module, it improves the power of the network to alleviate larger domain shift. Explicit discrepancy minimization can be achieved through the feature adaptation module at higher layers. In addition, we further propose a Generalized Domain Conditioned Adaptation Network (GDCAN) over DCAN by conditionally deciding whether the source and target domains enter different channel activation paths.
  This flexible adaptive routing strategy could facilitate more precise domain-informative knowledge extraction and transfer.
  
  To quantitatively measure domain discrepancy, we explore the standard distribution distance metric MMD~\cite{MMD,MMD_2012}, which can be formulated as follows:
  \begin{small}
  \begin{equation}\label{mmd}
  MMD(P_s, Q_t)=\sup_{h\in \mathcal{H}}\bigg\Vert\mathbb{E}_{\x_i^s \sim P_s} [h(\x_i^s)]
  -\mathbb{E}_{\x_j^t \sim Q_t} [h(\x_j^t)] \bigg\Vert^2,
  \end{equation}
  \end{small}%
  where $h$ is a non-linear feature map in Reproducing Kernel Hilbert Space (RKHS) $\mathcal{H}$. It has been proven that two distribution are equal if and only if $MMD(P_s,Q_t)=0$ \cite{MMD_2012}. The following describes the details of our approaches.

  \vspace{-3mm}
  \subsection{Domain-Specialized Feature Learning}\label{section:domain_specific_learning}

  Traditional deep DA schemes keep the network trained by source supervision unchanged to learn common features in source and target domains.
  However, in Fig. \ref{Fig_attention_visualization}(c), we show that a completely-shared convolutional network will falsely highlight irrelevant objects, which is similar to the attention maps of the source-only model. This is because the network will extract more source-relevant features rather than target-relevant features due to strong source supervision, hindering domain-specialized feature representation learning and resulting in target misclassification.
  
  \subsubsection{Domain Conditioned Channel Attention Module}
  To address the aforementioned problem, we introduce a domain conditioned channel attention module to facilitate feature recalibration in convolutional layers by preserving useful information while suppressing useless ones for each domain. It is a multi-path structure separating domains into different activation procedures, so as to model the independencies between the convolutional channels for source and target, respectively. In this way, domain-specialized features in low-level can be discovered and preserved, which further encourages feature alignment in task-specific layers. Meanwhile, we acknowledge that the deep network itself is able to extract features with powerful generalization abilities \cite{alexnet, resnet, densenet}. Thus, we allow these two paths to share partial model parameters due to the certain correlation between source and target domains. Motivated by this, we can guarantee that the target domain would perform feature recalibration to extract more domain-specialized feature descriptions in convolutional layers without losing adaptive information in the source domain.

  As shown in Fig.~\ref{Fig_method}, we denote the intermediate source and target feature embeddings as $\{\boldsymbol{X}_s, \boldsymbol{X}_t\} \in \mathbb{R}^{C\times H\times W}$, where $H$, $W$ are the spatial dimensions (height and width) and $C$ is the number of channels. Inspired by~\cite{senet}, we first generate channel descriptors $\boldsymbol{d} \in \mathbb{R}^{C\times 1\times 1}$ for each domain. We take global average pooling on $\{\boldsymbol{X}_s, \boldsymbol{X}_t\}$ for overall information extraction in each channel:
  \begin{small}
  \begin{equation}\label{avg-pooling}
  d^c=\frac{1}{H*W} \sum_{i=1}^{H}\sum_{j=1}^{W} \boldsymbol{X}_{ij}^c,
  \end{equation}
  \end{small}%
  where $d^c$ denotes the average value over all pixels of the $c^{th}$ channel, and $(i, j)$ means the location coordinate.

  To capture channel-wise dependencies in each domain, we partition the data stream $\boldsymbol{d} =\{\boldsymbol{d}_s, \boldsymbol{d}_t\}$ into two branches according to the domain label. As shown in Fig. \ref{Fig_method}, the blue and green arrows indicate target and source data flow, respectively. Each branch is followed with a dimensionality-reduction layer (i.e., fully-connected layer) with a ratio $\tau$\footnote{We fix $\tau=16$ in this paper as \cite{senet}.}. In this way, we can learn an interaction across channels and the intermediate channel descriptors are reshaped to $\frac{C}{\tau}\times 1\times 1$. With further ReLU activation \cite{relu}, we incorporate source and target streams together, and perform dimensionality increasing by forwarding them into the same FC-layer and rescaling function. Thus, the channel attention vectors are again reshaped from $1\times 1\times \frac{C}{\tau}$ to $1\times 1\times C$, as:
  \begin{small}
  \begin{equation}\label{source_weight}
  \boldsymbol{\omega_s}=\sigma\Big(FC(\mathrm{ReLU}({FC}_s(\boldsymbol{d}_s)))\Big),
  \end{equation}
  \begin{equation}\label{target_weight}
  \boldsymbol{\omega_t}=\sigma\Big(FC(\mathrm{ReLU}({FC}_t(\boldsymbol{d}_t)))\Big),
  \end{equation}
  \end{small}%
  where $\sigma (\cdot)=1/(1+e^{-x})$ is a Sigmoid function. Note that $FC(\cdot)$ denotes the shared FC-layer with the corresponding linear transformation for dimensionality increasing, while $FC_s(\cdot)$ and $FC_t(\cdot)$ are the separate dimensionality-reduction transformations for source and target domains. The attention weights $\boldsymbol{\omega_s}$ and $\boldsymbol{\omega_t}$ reflect the importance of channels across domains. As a result, the domain conditioned attention module decides how much attention paid to features at different channels for each domain.
  
  Then, we can obtain activated feature mappings by multiplying channel weights with the original features $\boldsymbol{X}_s$ and $\boldsymbol{X}_t$ channel-wisely, which are formulated as: 
  \begin{small}
    \begin{equation}\label{target_activated}
      \widetilde{\boldsymbol{X}}_s,=\boldsymbol{\omega_s}\odot \boldsymbol{X}_s, ~~~~ \widetilde{\boldsymbol{X}_t}=\boldsymbol{\omega_t}\odot \boldsymbol{X}_t,
      \end{equation}
  \end{small}%
  where $\odot$ denotes channel-wise multiplication and $\widetilde{\boldsymbol{X}}=\{\widetilde{\boldsymbol{X}_s},\widetilde{\boldsymbol{X}_t}\}$ are the recalibrated convolutional representations for source and target domains.
  
  In general, the attention module makes the target domain not only inherit the powerful feature extraction ability from the source network but also independently learn the importance of each feature channel, which will benefit the recalibration of target domain convolutional features. Due to its light-weight nature, the proposed channel attention module won't introduce many extra parameter costs. Besides, it can be easily incorporated into existing residual architecture of ResNet~\cite{resnet} and other popular DA methods~\cite{CDAN,MSTN,transnorm}.
  
  \subsubsection{Adaptive Channel Attention Module}\label{sec:adaptive_strategy}
  
  While domain conditioned attention module enhances the domain specificity of the learned representations, it enforces route separation in each module despite the high similarity of domain statistics in some layers.
  Thus, we further propose a Generalized Domain Conditioned Adaptation Network (GDCAN) with adaptive domain conditioned channel attention module that employs a strategy to make route decision on whether to separate domain channel activations in each module. As shown in Fig.~\ref{Fig_method}, we take one convolutional representation as an example. In this adaptive channel attention module, we first apply the adaptive routing strategy to determine the attention calculation path of the target domain based on the defined cross-domain statistic distance. If the distance is small enough, both domains share the attention computing routing (i.e., the source branch). Otherwise, source and target will proceed separately.
  
  Specifically, we calculate the mean and variance values of source and target intermediate convolutional representations, which reflect the information of feature distributions of both domains to some extent, and utilize them as domain statistics estimations:
  \begin{small}
  \begin{equation}\label{coefficient}
  m_s=\frac{\mu_s}{\sqrt{\sigma_s+\epsilon}},~~~ m_t=\frac{\mu_t}{\sqrt{\sigma_t+\epsilon}},
  \end{equation}
  \end{small}%
  where $\mu_s$ and $\mu_t$ are the means of source and target feature representations, respectively. $\sigma_s$ and $\sigma_t$ are their corresponding variances.
  $\epsilon$ is a small constant to avoid trivial division.
  Then, the cross-domain difference can be computed by the absolute value of the difference between $m_s$ and $m_t$.
  So far, we have not fully defined the statistic distance as the range of this difference is uncertain.
  To specify difference for normalization, we take target statistics $m_t$ as the relative value and restrict the normalized difference in $[0,1)$ by tanh function. The cross-domain statistic distance $\widehat{m}$ can be simply formulated as follows:
  \begin{small}
  \begin{equation}\label{distance}
  \widehat{m}=\tanh(\frac{\left|m_s-m_t\right|}{m_t}).
  \end{equation}
  \end{small}
  
  Since we map the statistic difference to a value in $[0,1)$, it is reasonable to set a threshold $\lambda$ to control the routing choice adaptively.
  If $\widehat{m}$ is smaller than $\lambda$, it means the source and target representations are much similar at this convolutional stage. Then target and source domains could pass through the identical source branch, sharing the source channel attention calculation weights; otherwise they use separate branches to derive domain-specific attentions, which can be formulated as:
  \begin{small}
  \begin{equation}
   \boldsymbol{\omega_t} =
    \begin{cases}
      \sigma(FC(\mathrm{ReLU}({FC}_s(\boldsymbol{d}_t)))), &\text{if $\widehat{m} < \lambda$}\\
    \sigma(FC(\mathrm{ReLU}({FC}_t(\boldsymbol{d}_t)))). &\text{otherwise.}
    \end{cases}
  \end{equation}
  \end{small}
  
  Compared with the static attention module in DCAN, the dynamic attention module in GDCAN applies an adaptive routing strategy to perform domain processing separation in a selective rather than compulsory way. Consequently, it can further improve the flexibility and ability in modeling complex data from different domains. 
  
  Actually, the selections of cross-domain statistic measures and the value of $\lambda$ are crucial for the proposed adaptive domain conditioned channel attention module. Thus, we have the following discussions.

  \noindent{\textbf{Discussion 1}}: \textit{How to measure the cross-domain statistic distance?}  \textbf{First of all}, the proposed statistic distance does not introduce extra parameters or modules. Different from~\cite{AutoDIAL}, we do not explicitly introduce new domain alignment layers that are embedded at different levels of the deep network. Instead, we leverage the statistical values of source and target intermediate convolutional representations themselves. \textbf{Second}, the proposed statistic distance can be treated as a simple and efficient surrogate. Precisely, in our framework, we could bring in any other metrics which can cope with the discrepancy across distributions. For example, in DA, MMD~\cite{MMD} has been, without doubt, one of the most applicable techniques to measure the cross-domain difference so far. Likewise, the KL-divergence as well can achieve this (see Section~\ref{adaptive_measurement}). However, they may bring a little extra computational cost when dealing with large-scale convolutional activations. \textbf{Third}, compared with~\cite{transnorm}, rather than calculating channel-wise distance per instance, we compute cross-domain static distance batch-wisely, which will benefit capturing the whole distribution information for both domains, and proceeding routing selection appropriately.

  \noindent{\textbf{Discussion 2}}: \textit{Intrigued, you are encouraged to think about how to decide the value of $\lambda$?} As disserted in~\cite{transferable}, knowledge transferability changes along with convolutional layers and various layers actually respond to different visual patterns. That is, the features learned by filters will evolve from low-level, such as lines and edges, to task-specific as the convolutional layers deepen. Thus, the transferability across layers varies, and it is encouraged to enforce the routing strategy adaptively. In this paper, for simplicity, we fix $\lambda = 0.2$ in all experiments (Section~\ref{sec:experiment}) to control the routing selection based on the cross-domain statistic distance. Moreover, we explore more flexible and dynamic $\lambda$ tuning strategies in Section~\ref{adaptive_measurement} to deeply analyze the effects of threshold choices.

  \noindent{\textbf{Discussion 3}}: \textit{The advantage of using domain conditioned channel attention for cross-channel domain alignment.} Roy et al.~\cite{DWT-MEC} whiten source and target features to a common spherical distribution, which is a generalization of BN. We argue that DWT aims to perform domain-specific standardization to the feature activations and handle the correlation among features, in which an extra normalization layer is involved after each convnets as moment matching between the source and target domains. Besides, there has been no effort on modeling channel-wise transferability in deep neural networks. Apparently, the transferability of each channel comprises a major obstruct in designing domain-specific architecture. Our work turns to channel attention for cross-channel domain alignment which can be effectively embedded within the deep networks. Also, the proposed domain conditioned channel attention in convolutional layers with partially-shared parameters to activate distinctly interested channels for each domain. This mechanism not only benefits the representation learning of inter-domain invariant features to reduce the inter-domain gap but also learns informative domain-specific features.
  
  \vspace{-2mm}
  \subsection{Domain-General Feature Learning}
  After obtaining domain-informative features in convolutional layers, we expect that domain-general features of high-level should also be effectively derived.
  To achieve this, the common strategy is to align the marginal distributions in the task-specific layers through distance metric, i.e., MMD, which is based on the proposition that transferability of features will decrease dramatically along the network \cite{transferable}.
  
  Unlike previous works \cite{RTN,DRCN}, we adapt all task-specific layers, and more importantly, explicitly measure domain discrepancy from a structural aspect.
  To be specific, we design a feature adaptation module and plug it into all higher task-specific layers, including the classification layer.
  This structure contains several adaptation layers with only target domain data passing through it during the alignment.
  We expect these additional layers to assist target domain in learning its distribution discrepancy with source domain, which ultimately contributes to reducing domain mismatch.
  
  In our architecture shown in Fig. \ref{Fig_method}, given the $l^{th}$ ($l \in \{1,2, \dots, L\}$, and $L$ denotes the number of the last layers) task-specific layer, we denote its outputs of source data and target data as $G_l({\boldsymbol{x}^s})$ (green feature vector) and $G_l({\boldsymbol{x}^t})$ (blue feature vector), respectively. Different from the source data flow $G_l({\boldsymbol{x}^s})$ directly into next task-specific layer, we additionally forward target embedding $G_l({\boldsymbol{x}^t})$ into a feature adaptation module consisting of $FC$, ReLU and $FC$ layers to generate a new feature vector $\Delta G_l({\boldsymbol{x}^t})$.
  Nevertheless, multi-layers mapping might lead to both optimization issue and severe degradation of useful information in original inputs.
  Therefore, we use skip-connection on target stream that bypasses linear transformations with the identity function: $\widehat{G_l}(\boldsymbol{x}^t)=G_l(\boldsymbol{x}^t)+\Delta G_l(\boldsymbol{x}^t)$, by which we can gain benefits from representative ability of $FC$ layers without significant information loss.
  In addition, we expect the added transformation $\Delta G_l({\boldsymbol{x}^t})$ to automatically capture the discrepancy between $G_l({\boldsymbol{x}^s})$ and $G_l({\boldsymbol{x}^t})$, thus making $\widehat{G_l}({\boldsymbol{x}^t}) \approx G_l({\boldsymbol{x}^s})$.
  For this reason, we align the marginal distributions between $G_l({\boldsymbol{x}^s})$ and $\widehat{G_l}({\boldsymbol{x}^t})$ via the classical MMD criterion \cite{DAN_pami}, which can be formulated as:
  \begin{small}
  \begin{equation}\label{f-mmd}
  \begin{aligned}
  \mathcal{L}_{\mathcal{M}}&= \sum_{l=1}^L MMD\big(G_l({\boldsymbol{X}_s}), \widehat{G_l}({\boldsymbol{X}_t})) \\
  &=\sum_{l=1}^L \bigg\Vert \frac{1}{n_s}\sum_{i=1}^{n_s}h \big(G_l({\boldsymbol{x}^s_i})\big)
  -\frac{1}{n_t}\sum_{j=1}^{n_t}h\big(\widehat{G_l}({\boldsymbol{x}^t_j})\big) \bigg\Vert^2_{\mathcal{H}},
  \end{aligned}
  \end{equation}
  \end{small}%
  where $\mathcal{L}_{\mathcal{M}}$ indicates the sum of MMD losses over all $L$ feature adaptation modules.
  By minimizing the Eq.~\eqref{f-mmd}, the distributions across two domains in each task-specific layer will be enforced in a shared embedding space, where their gap could be measured and reduced, accordingly, domain-invariant representations can be generated.
  Meantime, we consider building the adaptation module after the softmax layer as well, which facilitates transferring category correlation knowledge from source to target in a unified way.
  
  However, if naively conducting distribution matching, this global alignment strategy might lead to over-transferring between source and target, consequently destroying domain-wise structures while conveying noisy and non-essential information across domains.
  To further avoid the arbitrariness of adaptation learning, we enforce source data to pass through the adaptation module for regularization.
  Intuitively, the source domain representation should be unchanged after passing through the adaptation module, i.e., the distributions of $G_l(\boldsymbol{x}^s)$ and $\widehat{G_l}(\boldsymbol{x}^s)$ should keep similar.
  But if we exactly align each class in source domain, it will translate to $\Delta G_l(\boldsymbol{x}^s)\approx 0$. A possible consequence of such class-wise alignment is that adaptation module plays no role in mismatch reduction because it learns nothing.
  
  To address this problem, we propose a novel regularization loss that performs a compromising fashion. Specifically, it enforces a random size of source data passing through the target path, and minimizing its MMD metric with source domain. The regularization loss is formally written as:
  \begin{small}
  \begin{equation}\footnotesize\label{f-random}
  \begin{aligned}
  \mathcal{L}_{reg}&=\sum_{l=1}^L MMD\big(G_l({\boldsymbol{X}_s}), \widehat{G_l}({R}))\\
  &=\sum_{l=1}^L \sum_{k=1}^{\mathcal{C}_n}\bigg\Vert \frac{1}{n_s^{k}}\sum_{{\boldsymbol{x}^s_i}\in\mathcal{S}^{k}}h \big(G_l({\boldsymbol{x}^s_i})\big)
  -\frac{1}{|R|}\sum_{{\boldsymbol{x}^s_j}\in R}h\big(\widehat{G_l}({{\boldsymbol{x}^s_j}})\big) \bigg\Vert^2_{\mathcal{H}},
  \end{aligned}
  \end{equation}
  \end{small}%
  where $R$ is a random subset from source domain and $|R|$ is the set size which is stochastic. We select this subset by a probability option $\frac{p}{\mathcal{C}_n} \in [0, 1]$ with control factor $p$, which denotes the ratio of source samples allocated to the regularization set.
  This regularization term can not only appropriately guide the feature correction process, but also enhance the alignment ability of the adaptation module.
  More detailed ablation studies investigating the efficacy of key designs of $\mathcal{L}_{\mathcal{M}}$ and $\mathcal{L}_{reg}$ can be seen in Section \ref{sec:ablation_study}.

  \begin{figure*}[tb]
    \centering
    \includegraphics[width=0.95\textwidth]{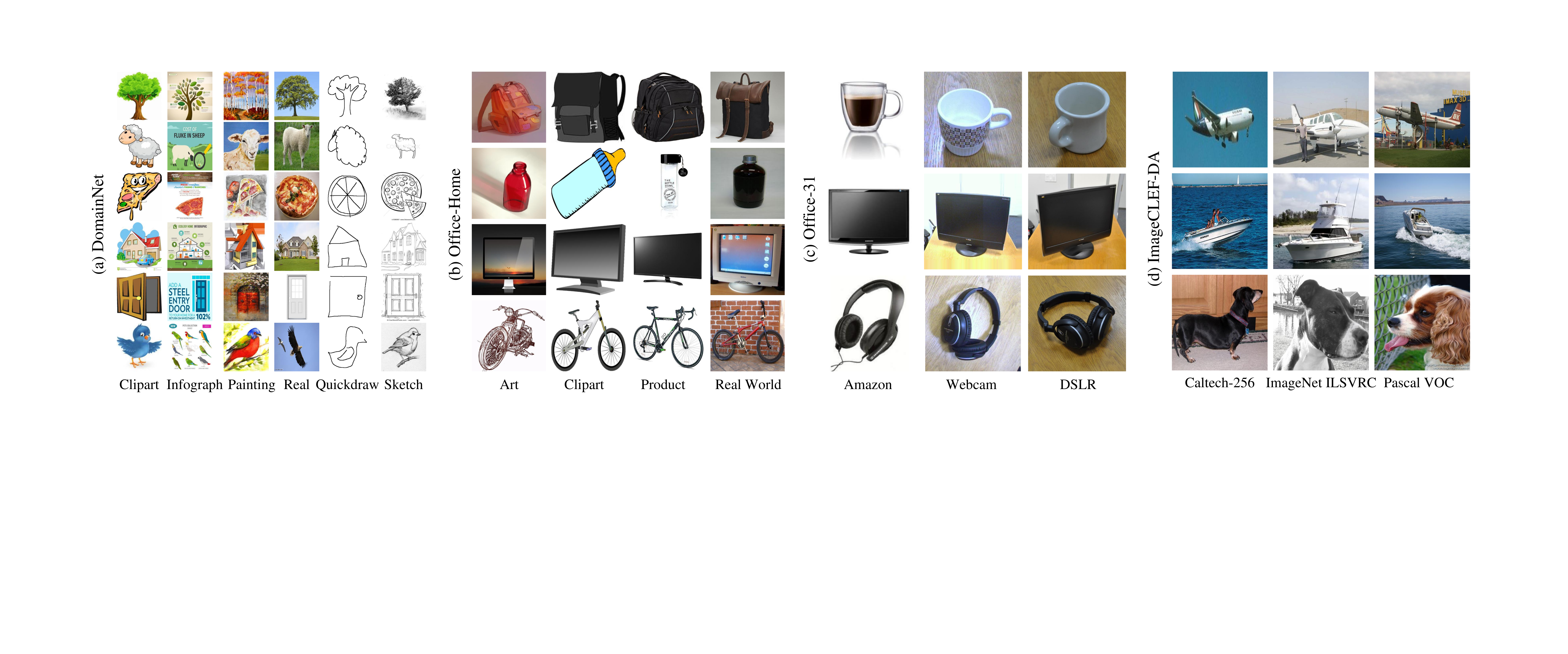}\vspace{-4mm}
    \caption{Example images from (a) DomainNet, (b) Office-Home, (c) Office-31, and (d) ImageCLEF-DA datasets.
     }
    \label{Fig_examples}\vspace{-4mm}
    \end{figure*}
    
  \vspace{-3mm}
  \subsection{Overall Objective}
  In the context of unsupervised domain adaptation, we follow the consensus that training process is performed on labeled source data and unlabeled target data. Thereafter, we can build a source classification loss under the supervision of ground-truth source label, which guarantees the discriminative learning of source domain. Mathematically, we minimize the following classification loss:
  \begin{small}
  \begin{equation}\label{f-source}
  \min_{F}~~~~\mathcal{L}_{s}=\frac{1}{n_s}\sum^{n_s}_{i=1}\mathcal{E}(F({\boldsymbol{x}^s_i}),{y^s_i}),
  \end{equation}
  \end{small}%
  where $\mathcal{E}(\cdot,\cdot)$ is the cross-entropy loss function and $F(\cdot)$ is the learned predictive model.
  
  On the other hand, this supervision is inclined to predict over-confidently on source-like data, which results in poor generalization performance on the target domain due to a lack of certainty in the prediction on target-like data.
  In this case, we adopt conditional entropy minimization \cite{MinEnt} of unlabeled target data, as used in \cite{SymNets,RTN,CDAN}.
  We define the $k^{th}$ class-conditional probability of target data ${\boldsymbol{x}^t}$ predicted by classifier as $F^{(k)}({\boldsymbol{x}^t})$. Then, the target entropy loss can be obtained as:
  \begin{small}
  \begin{equation}\label{entropy}
  \min_{F}~~~~\mathcal{L}_e=-\frac{1}{n_t}\sum_{j=1}^{n_t}\sum_{k=1}^{\mathcal{C}_n}F^{(k)}({\boldsymbol{x}^t_j})
  \mathrm{log}F^{(k)}({\boldsymbol{x}^t_j}).
  \end{equation}
  \end{small}%
  
  By minimizing Eq.~\eqref{entropy}, we can predict target data at a high level of confidence as well, which forces the classifier to pass through the low-density region of the target domain, and improves the classifier generalization ability.
  
  To summarize, the objective of our proposed method is to jointly optimize four components including task-specific feature alignment $\mathcal{L}_{\mathcal{M}}$, regularization loss $\mathcal{L}_{reg}$, source classification $\mathcal{L}_{s}$, and target entropy loss $\mathcal{L}_{e}$.
  The overall optimization problem can be written as follows:
  \begin{small}
  \begin{equation}\label{objective}
  \min_{F}~~~~\mathcal{L}=\mathcal{L}_s+\alpha(\mathcal{L}_\mathcal{M}+\mathcal{L}_{reg})+\beta\mathcal{L}_e,
  \end{equation}
  \end{small}%
  where the parameters $\alpha$ and $\beta$ weigh the relative importance of these loss terms. Here, DCAN and GDCAN provide a unified framework for DA problems, and more experimental comparisons can be seen in Section \ref{sec:experiment} and Section \ref{sec:analysis}.


\begin{table}[tb]\footnotesize
  \centering
  \caption{Statistics of the benchmark datasets.}\vspace{-4mm}
    \begin{tabular}{|c|ccr|c|}
    \hline
    Dataset & Sub-domain & Abbr. & \#Sample & \#Class \bigstrut\\
    \hline
    \multirow{6}[2]{*}{DomainNet} & Infograph & inf     & 51,605 & \multirow{6}[2]{*}{345} \bigstrut[t]\\
          & Quickdraw  & qdr     & 172,500 &  \\
          & Real  & rel     & 172,947 &  \\
          & Sketch  & skt     & 69,128 &  \\
          & Clipart  & clp     & 48,129 &  \\
          & Painting  & pnt     & 72,266 &  \\
    \hline
    \multirow{4}[2]{*}{Office-Home} & Art   & Ar    & 2,427  & \multirow{4}[2]{*}{65} \bigstrut[t]\\
          & Clipart & Cl    & 4,365  &  \\
          & Product & Pr    & 4,439  &  \\
          & Real-World & Rw    & 4,357  &  \\
    \hline
    \multirow{3}[1]{*}{Office-31} & Amazon & A     & 2,817  & \multirow{3}[1]{*}{31} \bigstrut[t]\\
          & DSLR  & D     & 498   &  \\
          & Webcam & W     & 795   &  \\
    \hline
    \multirow{3}[1]{*}{ImageCLEF-DA} & ImageNet & I     & 600  & \multirow{3}[1]{*}{12} \bigstrut[t]\\
          & Pascal  & P     & 600   &  \\
          & Caltech & C     & 600   &  \\
    \hline
    \end{tabular}%
  \label{tab:statistics-dataset}\vspace{-4mm}
\end{table}%

\section{Experiment}\label{sec:experiment}
In this section, we compare the proposed methods with several state-of-the-art unsupervised DA methods to demonstrate our effectiveness. In addition, we also validate the general applicability of the proposed DCAN and GDCAN across different variants as well as different architectures.

\subsection{Dataset}
We evaluate the proposed methods on four popular cross-domain image benchmarks: DomainNet \cite{DomainNet}, Office-Home \cite{Office-Home}, Office-31 \cite{Office31} and ImageCLEF-DA. Fig. \ref{Fig_examples} and Table \ref{tab:statistics-dataset} show example images in four benchmark datasets and their corresponding data statistics respectively.

\begin{table*}[!htbp]
  \caption{Accuracy(\%) on \textbf{DomainNet} for unsupervised DA. In each sub-table, the column-wise domains are selected as the source domain and the row-wise domains are selected as the target domain. ($\ddagger$ Implement according to the original source code.)}\vspace{-4mm}
  \centering
 \resizebox{\textwidth}{!}{
  \setlength{\tabcolsep}{0.5mm}{
    \begin{tabular}{c|ccccccc||c|ccccccclc|ccccccc||c|ccccccc}
    \multicolumn{16}{c}{Accuracy(\%) on DomainNet for UDA (\textbf{ResNet-50})}                                                &       & \multicolumn{16}{c}{Accuracy(\%) on DomainNet for UDA (\textbf{ResNet-101})} \\
    \toprule
    ResNet$^{\ddagger}$ &  clp  & inf    & pnt   & qdr   & rel   & skt   & Avg. & MCD$^{\ddagger}$ & clp   & inf   & pnt   & qdr   & rel   & skt   & Avg.  &       & ResNet & clp   & inf   & pnt   & qdr   & rel   & skt   & Avg.  &MCD  & clp   & inf   & pnt   & qdr   & rel   & skt   & Avg. \\
    \cline{1-16}\cline{18-33}  clp  & -  & 14.2 & 29.6 & 9.5 & 43.8 & 34.3 & 26.3 & clp   & - & 15.4  &  25.5 &  3.3 & 44.6  &  31.2 & 24.0  &    & clp   & -  & 19.3  & 37.5  & 11.1  & 52.2  & 41.0  & 32.2  & clp   & -     & 14.2  & 26.1  & 1.6   & 45.0  & 33.8  & 24.1 \\
    inf   & 21.8  & -  & 23.2 & 2.3   & 40.6  & 20.8  & 21.7  & inf   & 24.1 & - & 24.0  & 1.6   & 35.2  & 19.7  &  20.9 &       & inf   & 30.2  & -     & 31.2  & 3.6   & 44.0  & 27.9  & 27.4  & inf   & 23.6  & -     & 21.2  & 1.5   & 36.7  & 18.0  & 20.2 \\
    pnt   &  24.1 & 15.0 & -  & 4.6  &  45.0 &  29.0 &  23.5 & pnt & 31.1  & 14.8 & -  & 1.7 &  48.1 & 22.8 & 23.7  &       & pnt   & 39.6  & 18.7  & -     & 4.9   & 54.5  & 36.3  & 30.8  & pnt   & 34.4  & 14.8  & -     & 1.9   & 50.5  & 28.4  & 26.0 \\
    qdr   & 12.2 & 1.5   & 4.9 & -  & 5.6   & 5.7 & 6.0  & qdr   & 8.5 & 2.1 & 4.6 & -   & 7.9   & 7.1   & 6.0 &       & qdr   & 7.0   & 0.9   & 1.4   & -     & 4.1   & 8.3   & 4.3   & qdr   & 15.0  & 3.0   & 7.0   & -     & 11.5  & 10.2  & 9.3 \\
    rel   &  32.1 & 17.0  &  36.7 & 3.6   & -   & 26.2  & 23.1  & rel  &  39.4  & 17.8  &   41.2 & 1.5 & -  & 25.2  & 25.0 &   & rel   & 48.4  & 22.2  & 49.4  & 6.4   & -  & 38.8  & 33.0  & rel   & 42.6  & 19.6  & 42.6  & 2.2   & -     & 29.3  & 27.2 \\
    skt   &  30.4  & 11.3  & 27.8  & 3.4   & 32.9  & -  & 21.2  & skt  & 37.3 & 12.6  & 27.2  & 4.1   & 34.5  & -    & 23.1  &       & skt   & 46.9  & 15.4  & 37.0  & 10.9  & 47.0  & -     & 31.4  & skt   & 41.2  & 13.7  & 27.6  & 3.8   & 34.8  & -     & 24.2 \\
    Avg.  &  24.1 & 11.8 & 24.4 & 4.7 & 33.6  & 23.2  & 20.3 & Avg.  &  28.1 & 12.5 &  24.5 & 2.4 & 34.1 & 21.2 & 20.5 &       & Avg.  & 34.4  & 15.3  & 31.3  & 7.4   & 40.4  & 30.5  & 26.5  & Avg.  & 31.4  & 13.1  & 24.9  & 2.2   & 35.7  & 23.9  & 21.9 \\
    \cmidrule(lr){1-17}\cmidrule(lr){18-33}
    CDAN$^{\ddagger}$ & clp   & inf   & pnt   & qdr   & rel   & skt   & Avg.  & SWD$^{\ddagger}$ & clp   & inf   & pnt   & qdr   & rel   & skt   & Avg.  &       &DANN$^{\ddagger}$  & clp   & inf   & pnt   & qdr   & rel   & skt   & Avg.  & ADDA  & clp   & inf   & pnt   & qdr   & rel   & skt   & Avg. \\
    \cmidrule(lr){1-17}\cmidrule(lr){18-33}
    clp   & -  &  13.5 & 28.3 & 9.3 & 43.8 & 30.2 & 25.0  & clp   & -   & 14.7 & 31.9 & 10.1 & 45.3 & 36.5 & 27.7 & & clp & - & 14.8 & 32.7 & 12.3 & 48.3 & 34.2 & 28.4  & clp   & -     & 11.2  & 24.1  & 3.2   & 41.9  & 30.7  & 22.2 \\
    inf   &  18.9 & - & 21.4 & 1.9  & 36.3  & 21.3  & 20.0  & inf   & 22.9 & -  & 24.2 & 2.5 & 33.2 & 21.3 & 20.0 &  & inf   & 22.4 & -  & 25.9 & 2.8 & 35.2 & 19.8 & 21.2 & inf   & 19.1  & -     & 16.4  & 3.2   & 26.9  & 14.6  & 16.0 \\
    pnt  & 29.6  & 14.4 & - & 4.1 & 45.2 & 27.4 & 24.2 & pnt  & 33.6 & 15.3 & - & 4.4 & 46.1 & 30.7 & 26.0 &  & pnt & 34.1 & 14.9 & - & 4.9 & 48.4  & 31.0  & 26.7  & pnt   & 31.2  & 9.5   & -     & 8.4   & 39.1  & 25.4  & 22.7 \\
    qdr  & 11.8  & 1.2   & 4.0  & -  & 9.4   & 9.5 & 7.2 & qdr   & 15.5 & 2.2 & 6.4 & - & 11.1 & 10.2 & 9.1 &  & qdr  & 14.5 & 2.3 & 4.7  & - & 11.6 & 9.6 & 8.5 & qdr   & 15.7  & 2.6   & 5.4   & -     & 9.9   & 11.9  & 9.1 \\
    rel   & 36.4 & 18.3  & 40.9 & 3.4 & - & 24.6 & 24.7 & rel  & 41.2 & 18.1 & 44.2 & 4.6 & - & 31.6 & 27.9 &  & rel   & 40.6 & 16.4 & 43.1 & 5.3 & - & 30.2 & 27.1 & rel   & 39.5  & 14.5  & 29.1  & 12.1  & -     & 25.7  & 24.2 \\
    skt  & 38.2 & 14.7  & 33.9 & 7.0 & 36.6  & -  & 26.1 & skt & 44.2 & 15.2 & 37.3 & 10.3 & 44.7 & -  & 30.3  &       & skt   & 42.4 & 15.3 & 37.4 & 11.5  & 45.3  & -  &  30.4 & skt   & 35.3  & 8.9   & 25.2  & 14.9  & 37.6  & -     & 25.4 \\
    Avg.  & 27.0 & 12.4 & 25.7 & 5.1 & 34.3 & 22.6 & 21.2 & Avg.  & 31.5 & 13.1 & 28.8 & 6.4 & 36.1 & 26.1 & 23.6 &  & Avg.  & 30.8 & 12.7 & 28.7 & 7.4 & 37.8 & 24.9 & 23.7 & Avg.  & 28.2  & 9.3   & 20.1  & 8.4   & 31.1  & 21.7  & 19.8 \\

\cmidrule(lr){1-17}\cmidrule(lr){18-33}
    \multicolumn{1}{l|}{\textbf{DCAN}} & clp   & inf   & pnt   & qdr   & rel   & skt   & Avg.  & \multicolumn{1}{l|}{\textbf{GDCAN}} & clp   & inf   & pnt   & qdr   & rel   & skt   & Avg.  &       &\textbf{DCAN} & clp   & inf   & pnt   & qdr   & rel   & skt   & Avg.  &\textbf{GDCAN} & clp   & inf   & pnt   & qdr   & rel   & skt   & Avg. \\
    \cmidrule(lr){1-17}\cmidrule(lr){18-33}
    clp   & -  & 17.5 & 40.7 & 16.2 & 58.0 & 43.6 & 35.2 & clp & - & 18.2 & 41.9 & 16.5 & 58.7 & 44.0 & 35.9 & & clp & - & 18.5 & 43.6 & 17.1 & 60.3 & 45.8 & 37.1 & clp & - & 19.7 & 44.4 & 17.3 & 60.8 & 46.2 & 37.7 \\
    inf   & 35.0 & - & 34.8 & 3.8 & 24.4 & 27.4 & 25.1 & inf & 37.2 & - & 36.2 & 7.4 & 37.7 & 27.6 & 29.2 &  & inf   & 39.7 & -  & 38.4 & 5.9 & 54.6 & 28.5 & 33.4 & inf & 39.5  & -  & 39.2 & 9.1 & 55.1 & 31.4 & 34.9 \\
    pnt  & 46.3 & 18.5 & - & 8.3 & 60.2 & 38.8 & 34.4 & pnt  & 47.8 & 19.1 & - & 9.4 & 61.0 & 39.6 & 35.4 &  & pnt & 48.6 & 19.7 & - & 9.9 & 61.7 & 41.2 & 36.2 & pnt & 49.7 & 20.4 & - &  10.1 & 62.8 & 42.7 & 37.1 \\
    qdr  & 30.0 &  3.7 & 14.2 & -  & 14.5 & 12.3 & 14.9 & qdr & 31.3 & 6.4 & 14.6 & - & 25.1 & 20.9 & 19.7 &  & qdr  & 33.2 & 5.6 & 16.1 & - & 18.4 & 16.2 & 17.9 & qdr & 33.8 & 8.0 & 17.4 & - & 28.5 & 24.1 & 22.4 \\
    rel  & 50.9 & 17.5 & 48.1 & 2.7 & - & 31.1 & 30.1 & rel  & 52.3 & 20.4 & 48.5 & 9.8 & - & 37.6 & 33.7 &  & rel   & 53.7 & 18.5 & 50.5 & 4.0 & - & 33.4 & 32.0 & rel & 54.1 &  21.8 & 50.7 & 10.8 & - & 40.8 & 35.6 \\
    skt  & 55.2 & 16.2 & 44.6 & 8.7 & 53.2 & -  & 35.6 & skt & 55.8 & 18.6 & 46.7 & 16.7 & 57.8 & -  & 39.1 &  & skt & 57.6 & 17.3 & 47.3 & 10.1 & 55.3 & - & 37.5 & skt & 58.3 & 19.9 & 47.9 & 17.7 & 60.0 & - & 40.8 \\
    Avg. & 43.5 & 14.7 & 36.5 & 7.9 & 42.1 & 35.4 & \underline{29.2} & Avg. & 44.9 & 16.5 & 37.6 & 12.0 & 48.1 & 33.9 & \textbf{32.2} & & Avg. & 46.6 & 15.9 & 39.2 & 9.4 & 50.1 & 33.0 & \underline{32.4} & Avg. & 47.1 & 18.0 & 39.9 & 13.0 & 53.4 & 37.0 & \textbf{34.7} \\
    \bottomrule
    \end{tabular}
    }
    }\vspace{-4mm}
\label{tab:domainnet}
\end{table*}

\textbf{DomainNet} \cite{DomainNet}
is currently the largest DA image benchmark, which contains about 590k images of 345 categories in six domains. We refer these domains as: Infograph (inf), Quickdraw (qdr), Real (rel), Sketch (skt), Clipart (clp), Painting (pnt). Each domain has training set and test set without overlap. Additionally, this dataset has complex objects and scenes, e.g., furniture, mammal, building, etc. The large variations in pose, resolution, and modalities across domains are apparent in Fig. \ref{Fig_examples}, making DA extremely challenging under DomainNet. Following \cite{DomainNet}, we build 30 transfer tasks: inf $\rightarrow$ qdr, ..., pnt $\rightarrow$ clp, and notably, only training sets of both domains are involved in the training procedure, and the results of the target test set are reported.

\textbf{Office-Home} \cite{Office-Home}
is a popular dataset in office and home environments with nearly 15,600 samples of 65 categories.
It contains four distinct domains: Artistic (Ar), Clip Art (Cl), Product (Pr), and Real-World (Rw).
Specifically, Ar are images from paintings, sketches, or artistic depictions, and Cl are clipart pictures.
Pr are images without background and Rw is collected by cameras.
As a result, a total of 12 tasks will be available: Ar $\rightarrow$ Cl, ..., Rw $\rightarrow$ Pr.

\textbf{Office-31} \cite{Office31}
is a widely used dataset for DA, including over 4,000 images. Those images are divided into three domains: Amazon (A), DSLR (D), and Webcam (W). Each domain contains 31 categories found in the office setting, such as laptops, keyboards, backpacks.
Apparently, the size of Office-31 is much smaller and the tasks are easier than DomainNet and Office-Home.
As \cite{MDD}, we construct 6 cross-domain tasks: A $\rightarrow$ W, ..., W $\rightarrow$ A, D $\rightarrow$ A.

\textbf{ImageCLEF-DA}
is a standard benchmark for ImageCLEF 2014 domain adaptation challenge\footnote{http://imageclef.org/2014/adaptation}.
It is composed of 12 classes shared by four public datasets and denotes each dataset as a domain, which includes Caltech-256 (C), ImageNet ILSVRC 2012 (I), Pascal VOC 2012 (P), and Bing (B).
Different from the above datasets, it is number-balanced with 600 images in each domain and 50 images in each class.
As \cite{DAN_pami}, we consider three domains combinations (i.e., C, I, P), and thus build six cross-domain tasks: I $\rightarrow$ P, ..., P $\rightarrow$ C.

\begin{table*}[!htbp] \footnotesize
 \centering
  \caption{Accuracy (\%) on \textbf{Office-Home} for unsupervised DA (ResNet-50).}\vspace{-4mm}
  \setlength{\tabcolsep}{1.0mm}{
    \begin{tabular}{cccccccccccccc}
    \toprule
    Methods & Ar $\rightarrow$ Cl &Ar $\rightarrow$ Pr &Ar $\rightarrow$ Rw &Cl $\rightarrow$ Ar &Cl $\rightarrow$ Pr &Cl $\rightarrow$ Rw &Pr $\rightarrow$ Ar &Pr $\rightarrow$ Cl &Pr $\rightarrow$ Rw &Rw $\rightarrow$ Ar &Rw $\rightarrow$ Cl &Rw $\rightarrow$ Pr &Avg. \\
    \hline
    ResNet & 34.9  & 50.0  & 58.0  & 37.4  & 41.9  & 46.2  & 38.5  & 31.2  & 60.4  & 53.9  & 41.2  & 59.9  & 46.1  \\
    SRM & 47.3 & 68.1 & 77.1 & 45.5 & 60.6 & 63.3 & 50.0 & 42.5 & 75.5 & 65.6 & 46.2 & 78.5 & 60.0 \\
    DAN   & 43.6  & 57.0  & 67.9  & 45.8  & 56.5  & 60.4  & 44.0  & 43.6  & 67.7  & 63.1  & 51.5  & 74.3  & 56.3  \\
    DANN & 45.6  & 59.3  & 70.1  & 47.0  & 58.5  & 60.9  & 46.1  & 43.7  & 68.5  & 63.2  & 51.8  & 76.8  & 57.6  \\
    JAN   & 45.9  & 61.2  & 68.9  & 50.4  & 59.7  & 61.0  & 45.8  & 43.4  & 70.3  & 63.9  & 52.4  & 76.8  & 58.3  \\
    DWT & 50.3  & 72.1  & 77.0  & 59.6  & 69.3  & 70.2  & 58.3  & 48.1  & 77.3  & 69.3  & 53.6  & 82.0  & 65.6  \\
    CDAN & 50.7  & 70.6  & 76.0  & 57.6  & 70.0  & 70.0  & 57.4  & 50.9  & 77.3  & 70.9  & 56.7  & 81.6  & 65.8  \\
    TADA  & 53.1  & 72.3  & 77.2  & 59.1  & 71.2  & 72.1  & 59.7  & 53.1  & 78.4  & 72.4  & 60.0  & 82.9  & 67.6  \\
    SymNets & 47.7  & 72.9  & 78.5  & 64.2  & 71.3  & 74.2  & 64.2  & 48.8  & 79.5  & \textbf{74.5} & 52.6  & 82.7  & 67.6  \\
    TransNorm & 50.2 & 71.4 & 77.4 & 59.3 & 72.7 & 73.1 & 61.0 & 53.1 & 79.5 & 71.9 & 59.0 & 82.9 & 67.6 \\
    MDD   & 54.9 & 73.7  & 77.8  & 60.0  & 71.4  & 71.8  & 61.2  & 53.6 & 78.1  & 72.5  & 60.2 & 82.3  & 68.1  \\
    SAFN &  54.4 & 73.3 & 77.9 & 65.2 & 71.5 & 73.2 & 63.6 & 52.6 & 78.2 & 72.3 & 58.0 & 82.1 & 68.5 \\
    \hline
    \textbf{DCAN}  & 54.5  & \textbf{75.7} & 81.2 & 67.4 & \textbf{74.0} & 76.3 & \textbf{67.4} & 52.7  & 80.6 & 74.1  & 59.1  & \textbf{83.5} & 70.5 \\
    \textbf{GDCAN} & \textbf{57.3}  & \textbf{75.7}  & \textbf{83.1}  & \textbf{68.6}  & 73.2  & \textbf{77.3}  & 66.7  & \textbf{56.4}  & \textbf{82.2}  & 74.1  & \textbf{60.7}  & 83.0  & \textbf{71.5} \\
    \bottomrule
    \end{tabular}%
    }\vspace{-4mm}
  \label{tab:office-home}
\end{table*}

\subsection{Setup}
We implement all the methods in PyTorch~\cite{paszke2019pytorch} and use ResNet~\cite{resnet}, ResNext~\cite{ResNext} pre-trained on ImageNet~\cite{imagenet-dataset} as the backbone networks. Thus, the value L of task-specific layers is 2 (including the average pooling layer and softmax layer).
For a fair comparison, deep DA methods in our paper are under the standard unsupervised domain adaptation experiment settings \cite{JAN, DA_bp, MCD}.
All images are normalized to 256 $\times$ 256 and then randomly cropped to 224 $\times$ 224 as the network input.
We evaluate each transfer task using three random experiments.
In addition, we adopt stochastic gradient descent (SGD) with momentum of 0.9 and the learning rate annealing strategy as described in~\cite{DANN}.
In the experiments, we use a small batch of 36 samples per domain, therefore we freeze the BN layers~\cite{BN} and only update the weights of other layers through back-propagation.
Since the classification layer is trained from scratch, we set its learning rate to 10 times that of the other layers. By contrast, the learning rate of adaptation modules is 1/10 times because of its precision.
The hyper-parameter $p$ for adaption module is selected from the set $\{0.2, 0.4, 0.6, 0.8, 1\}$ according to the importance weighted cross-validation method as \cite{MDD}. We set coefficients $\alpha=1.5$ and $\beta=0.1$ throughout the paper, and parameter sensitivity analysis experiments are conducted to verify our methods could perform stably under the parameter varying. Since some papers follow the same experimental set-up like ours, we report their results in the published papers directly. Others are obtained by running their available source codes.

\begin{table*}[!htbp] \footnotesize
  \centering
  \caption{Accuracy (\%) on \textbf{Office-31} for unsupervised DA (ResNet-50).}\vspace{-4mm}
  \setlength{\tabcolsep}{0.5mm}{
    \begin{tabular}{ccccccccccccccccc|cc}
    \toprule
    Tasks    & ResNet & SRM & DAN  & RTN  & DANN & ADDA  & GTA  & DAAA & SAFN & CDAN  & DSBN    & TADA & SymNets & MDD & SPCAN & TransNorm & \textbf{DCAN} &\textbf{GDCAN}   \\
    \hline
    A $\rightarrow$ W  & 68.4 & 69.6 & 80.5 & 84.5 & 82.0 & 86.2 & 89.5 & 86.8 & 90.3 & 94.1 & 92.7      & 94.3 & 90.8    & 94.5   & 92.4  & \textbf{95.7}  & 95.0  & 94.8 \\
    D $\rightarrow$ W  & 96.7 & 97.3 & 97.1 & 96.8 & 96.9 & 96.2 & 97.9 & \textbf{99.3} & 98.7 & 98.6   & 99.0           & 98.7 & 98.8       & 98.4    & 99.2 & 98.7 & 97.5     &98.2      \\
    W $\rightarrow$ D  & 99.3 & 100.0 & 99.6 & 99.4 & 99.1 & 98.4 & 99.8 & \textbf{100.0} & \textbf{100.0} & \textbf{100.0} & \textbf{100.0} & 99.8 & \textbf{100.0} & \textbf{100.0} & \textbf{100.0} & \textbf{100.0} & \textbf{100.0} & \textbf{100.0} \\
    A $\rightarrow$ D  & 68.9 & 78.4 & 78.6 & 77.5 & 79.7 & 77.8 & 87.7 & 88.8 & 92.9 &  92.1 & 92.2  & 91.6 & 93.9  & 93.5  & 91.2 & \textbf{94.0} & 92.6  & 93.6     \\
    D $\rightarrow$ A  & 62.5 & 64.8 & 63.6 & 66.2 & 68.2 & 69.5 & 72.8 & 74.3  & 73.4 & 71.0           & 71.7   & 72.9 & 74.6     & 74.6    & 77.1 & 73.4 & \textbf{77.2}   & 76.9\\
    W $\rightarrow$ A  & 60.7 & 64.2 & 62.8 & 64.8 & 67.4 & 68.9 & 71.4 & 73.9  & 71.2 & 69.3           & 74.4   & 73.0 & 72.5 &  72.2  & 74.5 & 74.2  & \textbf{74.9}  &  74.4        \\
    Avg.      & 76.1 & 79.1 & 80.4 & 81.6 & 82.2 & 82.9 & 86.5 & 87.2 & 87.6 & 87.7 & 88.3      & 88.4 & 88.4  &  88.9    & 89.1  & 89.3 &  89.5    &  \textbf{89.7}    \\
    \bottomrule
    \end{tabular}%
    }\vspace{-4mm}
  \label{tab:office-31}
\end{table*}

\begin{figure*}[tb]
  \centering
  \includegraphics[width=\textwidth]{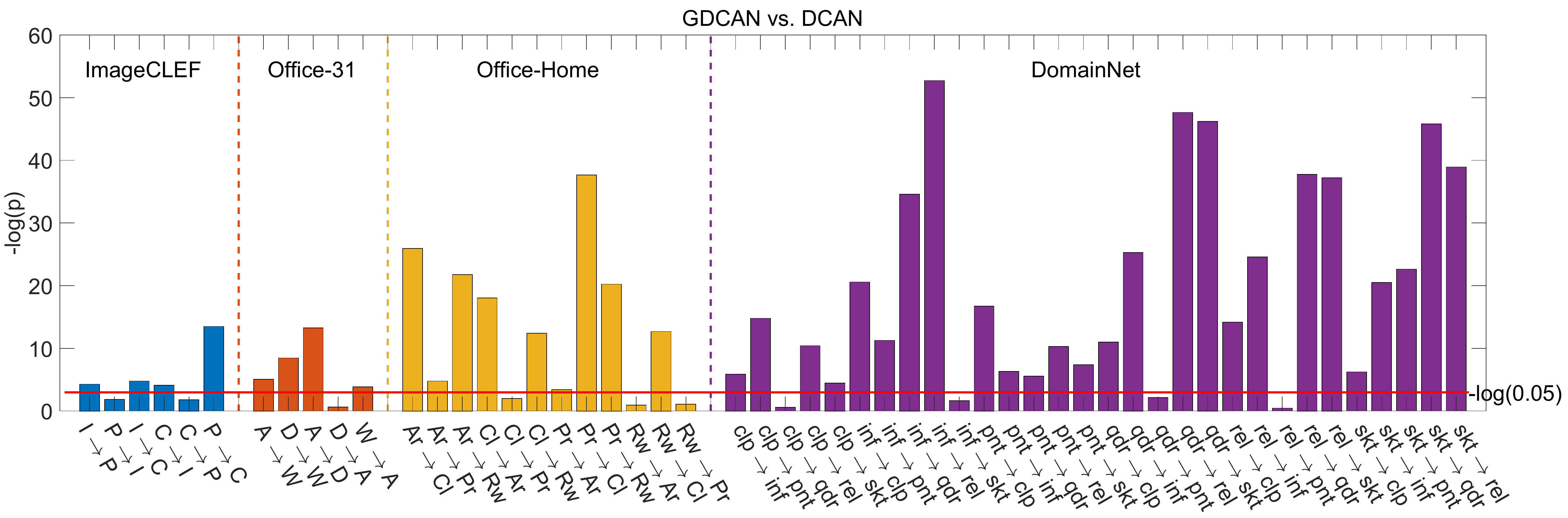}\vspace{-4mm}
  \caption{p-value of the significance test (t-test) for results of GDCAN vs. DCAN on all transfer tasks. To clearly illustrate the statistical significance, the base significance level of 0.05 ($-\log(0.05)$) is shown in red line. The larger value of $-\log(p)$ means the more significance of GDCAN.}\vspace{-4mm}
  \label{t_test}
  \end{figure*}

\subsection{Results of DomainNet}\label{sec:exp_domainnet}
As reported in Table~\ref{tab:domainnet}, we try different backbone networks and accordingly evaluate DCAN and GDCAN on the most challenging DA dataset DomainNet. We can observe that GDCAN significantly improves the accuracy on most tasks either in ResNet-50 backbone or ResNet-101 backbone, outperforming other baseline methods by a large margin.

With the ResNet-50 based network, DCAN and GDCAN bring $\mathbf{8.9\%}$ and $\mathbf{11.9\%}$ average accuracy improvements compared to the source-only model. In particular, GDCAN achieves new state-of-the-art results on DomainNet, surpassing SWD with a $\mathbf{8.6\%}$ margin.

Likewise, ResNet-101 based GDCAN obtains the highest average accuracy of $\mathbf{34.7\%}$ followed by ResNet-101 based DCAN with $\mathbf{32.4\%}$, while the accuracy is only $\mathbf{26.5\%}$ without conducting adaptation.
This implies that under a stronger backbone, our method can still substantially promote classification accuracies.
Note that negative transfer \cite{survey} occurs in some cases, where DA methods perform worse than source-only models.
We believe that it is due to the challenge of large class variation in DomianNet, which greatly increases the difficulty of tasks.
However, the overall performance of our method is still superior to others, which substantiates our work is suitable for very large-scale domain adaptation. Ultimately, we draw a conclusion that GDCAN can enrich low-level domain-specialized information to help learn more transferable features on this challenging dataset with the huge shift.

\subsection{Results of Office-Home}\label{sec:exp_officehome}
In Table~\ref{tab:office-home}, we summarize classification accuracies on Office-Home dataset.
Our proposed GDCAN brings the improvements up to $\mathbf{3.0\%}$ over the best baseline SAFN and surpass other baseline methods in almost twelve transfer tasks.
The considerable improvement reflects the designed modules successfully serve their functions in extracting two different characteristics of data.
More importantly, GDCAN achieves the best prediction accuracy of $\mathbf{71.4\%}$, by a margin of $\mathbf{1\%}$ over DCAN, which demonstrates that the adaptive attention module in GDCAN can better capture domain-specific knowledge across different domains. 

\subsection{Results of Office-31}\label{sec:exp_office31}
The experiment results on Office-31 dataset are shown in Table \ref{tab:office-31}.
A clear discovery is that our work yields better performance in hard tasks in which the difference between domains is significant, and produces comparable results in easy tasks.
For instance, we achieve top-1 accuracy on tasks D $\rightarrow$ A and W $\rightarrow$ A, whereas DAAA and TransNorm win the first place on tasks D $\rightarrow$ W and A $\rightarrow$ D, respectively.
This is an expected result since our domain-specialized feature learning may not further enhance the performance when two domains are much similar (i.e., over $90\%$ accuracies).
It can be also confirmed from the relatively small accuracy improvement from DCAN to GDCAN. This result suggests that the adaptive attention module may have limited capacity in easy transfer tasks due to the smaller domain gap. It is also consistent with our goal that attention module aims to extract feature that is domain-specific for each domain.

\begin{table}[htbp]\footnotesize
  \centering
  \caption{Accuracy (\%) on \textbf{ImageCLEF-DA} for unsupervised DA (ResNet-50).} \vspace{-4mm}
  \setlength{\tabcolsep}{1.0mm}{
    \begin{tabular}{cccccccc}
    \toprule
    Method & I $\rightarrow$ P    & P $\rightarrow$ I    & I $\rightarrow$ C    & C $\rightarrow$ I    & C $\rightarrow$ P    & P $\rightarrow$ C    & Avg. \\
    \hline
    ResNet  & 74.8  & 83.9  & 91.5  & 78.0  & 65.5  & 91.2  & 80.8  \\
    SRM & 77.0 & 89.3 & 94.7 & 84.8 & 70.5 & 93.5 & 85.0 \\
    DAN    & 74.5  & 82.2  & 92.8  & 86.3  & 69.2  & 89.8  & 82.5  \\
    RTN    & 75.6  & 86.8  & 95.3  & 86.9  & 72.7  & 92.2  & 84.9  \\
    DANN  & 75.0  & 86.0  & 96.2  & 87.0  & 74.3  & 91.5  & 85.0  \\
    JAN    & 76.8  & 88.0  & 94.7  & 89.5  & 74.2  & 91.7  & 85.8  \\
    MADA   & 75.0  & 87.9  & 96.0  & 88.8  & 75.2  & 92.2  & 85.9  \\
    CDAN   & 76.7  & 90.6   & \textbf{ 97.0 } & 90.5 & 74.5  & 93.5  & 87.1  \\
    SPCAN  & 79.0  & 91.1 & 95.5 & 92.9 & \textbf{79.4} & 91.3 & 88.2 \\
    TransNorm & 78.3 & 90.8 & 96.7 & \textbf{92.3} & 78.0 & 94.8 & 88.5 \\
    \hline
    \textbf{DCAN} & 80.5 & 91.2  & 95.7  & 91.8  & 77.2 & 93.3 & 88.3 \\
    \textbf{GDCAN} & \textbf{80.8} & \textbf{91.3} & 96.3  & 91.0  & 77.5 & \textbf{95.0} & \textbf{88.7} \\
    \bottomrule
    \end{tabular}%
    }\vspace{-4mm}
  \label{tab:imageclef}%
\end{table}%

\subsection{Results of ImageCLEF-DA}\label{sec:exp_imageclef}
Similar to the analysis of Office-31, GDCAN and DCAN achieve comparable results.
The average classification accuracy of GDCAN is $\mathbf{88.7\%}$ and that of DCAN is $\mathbf{88.3\%}$, which are $\mathbf{1.4\%}$ and $\mathbf{0.9\%}$ higher than CDAN, respectively.
We can see that the performance gain is greater on hard tasks, and less on easy tasks. For the task with an accuracy over $90\%$, our work gets slightly higher results than CDAN, such as tasks P $\rightarrow$ I, C $\rightarrow$ I and P $\rightarrow$ C. Although CDAN has the highest accuracy in the task I $\rightarrow$ C, GDCAN outperforms CDAN by a margin of $\mathbf{4.1\%}$ and $\mathbf{3.0\%}$ in hard tasks I $\rightarrow$ P and C $\rightarrow$ P. And this further validates that GDCAN is effective in modeling more complex representations when domain invariant transfer is limited.

\begin{table*}[htbp] \footnotesize
  \centering
  \caption{Accuracy (\%) on \textbf{Office-Home} for unsupervised DA (DCAN/GDCAN as the incremental module applied on different DA methods and CNNs).}\vspace{-4mm}
  \setlength{\tabcolsep}{1.0mm}{
    \begin{tabular}{lccccccccccccc}
    \toprule
    Method  & Ar $\rightarrow$ Cl  & Ar $\rightarrow$ Pr  & Ar $\rightarrow$ Rw  & Cl $\rightarrow$ Ar  & Cl $\rightarrow$ Pr  & Cl $\rightarrow$ Rw  & Pr $\rightarrow$ Ar  & Pr $\rightarrow$ Cl  & Pr $\rightarrow$ Rw  & Rw $\rightarrow$ Ar  & Rw $\rightarrow$ Cl  & Rw $\rightarrow$ Pr  & Avg. \\
    \hline
    MSTN  & 45.1  & 63.6   & 71.0  & 50.4  & 62.6  &  63.1  & 49.0  & 47.2  & 71.5  & 64.6 & 54.5  & 79.5 & 60.2 \bigstrut[t]\\
    + DCAN & 53.3  & 69.0  & 77.8  & 60.2  & 70.1  & 70.3  & 59.8  & 51.9  & 78.2   & 70.5   & 58.5  & 81.7  & 66.8 \\
    + GDCAN & \textbf{54.8} & \textbf{69.1}  & \textbf{78.1}  & \textbf{61.7}  & \textbf{71.5}  & \textbf{71.8}  & \textbf{61.3}  & \textbf{53.2}  & \textbf{79.1}   & \textbf{71.2} & \textbf{59.2}  & \textbf{82.3}  & \textbf{67.8}  \\
    \hline
    CDAN  & 49.0  & 69.3  & 74.5  & 54.4  & 66.0  & 68.4  & 55.6  & 48.3  & 75.9  & 68.4  & 55.4  & 80.5  & 63.8  \bigstrut[t]\\
    + DCAN & 54.8  & 74.2  & 80.9  & 65.6  & 72.8  & 76.2  & 64.1  & 52.5  & 81.7  & 71.4  & 57.9  & 83.6  & 69.6  \\
    + GDCAN & \textbf{55.1}   & \textbf{74.7}  & \textbf{81.6}  & \textbf{66.2}   & \textbf{73.3}  & \textbf{76.8}  & \textbf{64.8}  & \textbf{52.9}  & \textbf{82.4}  & \textbf{71.9}  & \textbf{58.6}  & \textbf{84.9}  & \textbf{70.3} \\
    \hline
    TransNorm & 50.2 & 71.4 & 77.4 & 59.3 & 72.7 & 73.1 & 61.0 & 53.1 & 79.5 & 71.9 & 59.0 & 82.9 & 67.6 \\
    + DCAN & 55.6 & 74.7 & 83.3 & 68.9 & 77.0 & 77.1 & 66.8 & 55.3 & 82.5 & 74.1 & 62.1 & 85.3 & 72.1 \\
    + GDCAN & \textbf{59.8} & \textbf{76.1} & \textbf{84.0} & \textbf{73.2} & \textbf{77.9} & \textbf{80.4} & \textbf{68.8} & \textbf{57.5} & \textbf{83.5} & \textbf{75.6} & \textbf{62.8} & \textbf{86.9} & \textbf{73.9} \\
    \hline
    ResNet-50 & 34.9  & 50.0  & 58.0  & 37.4  & 41.9  & 46.2  & 38.5  & 31.2  & 60.4  & 53.9  & 41.2  & 59.9  & 46.1  \bigstrut[t]\\
    + DCAN & 54.5  & \textbf{75.7}  & 81.2  & 67.4  & \textbf{74.0}  & 76.3  & \textbf{67.4} & 52.7  & 80.6  & \textbf{74.1}  & 59.1  & \textbf{83.5}  & 70.5 \\
    + GDCAN & \textbf{57.3}  & \textbf{75.7}  & \textbf{83.1}  & \textbf{68.6}  & 73.2  & \textbf{77.3}  & 66.7  & \textbf{56.4}  & \textbf{82.2}  & \textbf{74.1}  & \textbf{60.7}  & 83.0  & \textbf{71.5} \\
    \hline
    ResNext-50 & 41.1  & 65.5  & 74.5  & 53.0  & 63.7  & 66.3  & 51.6  & 37.6  & 72.7  & 62.4  & 41.3  & 74.3  & 58.7  \bigstrut[t]\\
    + DCAN & 58.6  & 75.1  & 83.3  & 72.3  & 76.1  & 78.7  & \textbf{71.3}  & \textbf{57.8}  & 83.5  & 75.7  & 60.1  & 83.7  & 73.0 \\
    + GDCAN & \textbf{59.5}  & \textbf{75.7}  & \textbf{83.8}  & \textbf{72.5}  & \textbf{77.3}  & \textbf{79.6}  & 70.4  & 57.2  & \textbf{83.6} & \textbf{76.6}  & \textbf{60.5}  & \textbf{84.5}  & \textbf{73.4}  \\
     \bottomrule
    \end{tabular}\vspace{-4mm}
  \label{tab:increment1}%
  }
\end{table*}%

\begin{table}[htbp]\footnotesize
  \centering
  \caption{Accuracy (\%) on \textbf{Office-31} for unsupervised DA (DCAN/GDCAN as the incremental module applied on different DA methods and CNNs).}\vspace{-4mm}
   \setlength{\tabcolsep}{0.6mm}{
    \begin{tabular}{lccccccc}
     \toprule
    Method & A $\rightarrow$ W    & D $\rightarrow$ W    & W $\rightarrow$ D    & A $\rightarrow$ D    & D $\rightarrow$ A    & W $\rightarrow$ A    & Avg. \\
     \hline
    MSTN  & 91.3 & 98.9 & \textbf{100.0}  & 90.4  & 72.7  & 65.6  & 86.5 \bigstrut[t]\\
    + DCAN & 92.8   & 97.3  & \textbf{100.0}  & 90.9  & 73.7 & 71.0  & 87.6 \\
    + GDCAN & \textbf{93.3}  & \textbf{97.9}  & \textbf{100.0}  & \textbf{91.2}  & \textbf{74.2}  & \textbf{72.2}  & \textbf{88.1} \\
     \hline
    CDAN  & \textbf{94.1}  & 98.6  & \textbf{100.0}   & 92.9  & 71.0    & 69.3  & 87.7 \bigstrut[t]\\
    + DCAN & 92.7  & \textbf{98.9}  & \textbf{100.0}   & 93.6  & \textbf{75.9}  & 72.7  & 89.0 \\
    + GDCAN & 93.3   & 98.5  & \textbf{100.0} & \textbf{94.0}  &  75.5 & \textbf{74.2}  & \textbf{89.3} \\
    \hline
    TransNorm  & 95.7  & 98.7  & \textbf{100.0}   & 94.0  & 73.4    & 74.2  & 89.3 \bigstrut[t]\\
    + DCAN & 95.8  & 98.7  & \textbf{100.0}   & 94.6  & 77.0  & \textbf{76.1}  & 90.3 \\
    + GDCAN & \textbf{96.5}   & \textbf{98.8}  & \textbf{100.0} & \textbf{95.9}  &  \textbf{78.1} & 76.0  & \textbf{90.8} \\
     \hline
    ResNet-50 & 68.4  & 96.7  & 99.3  & 68.9  & 62.5  & 60.7  & 76.1 \bigstrut[t]\\
    + DCAN & \textbf{95.0}    & 97.5  & \textbf{100.0}   & 92.6  & \textbf{77.2}  & \textbf{74.9}  & 89.5\\
    + GDCAN & 94.8    & \textbf{98.2}  & \textbf{100.0}   & \textbf{93.6}  & 76.9  & 74.4  & \textbf{89.7} \\
    \hline
    ResNext-50 & 76.1  & 96.2  & \textbf{99.8}  & 80.3  & 68.2  & 68    & 81.4 \bigstrut[t]\\
    + DCAN & 94.6  & \textbf{98.0}   & 99.6  & 95.0    & \textbf{77.3}  & 76.3  & 90.1 \\
    + GDCAN & \textbf{94.8}  & 97.7  & 99.2  & \textbf{95.6}  & 77.2  & \textbf{77.2}  & \textbf{90.3} \\
    \bottomrule
    \end{tabular}%
    }\vspace{-4mm}
  \label{tab:increment2}%
\end{table}%

\subsection{Significance Test (t-test)}
To further verify the effectiveness of the adaptive version of domain conditioned channel attention for all transfer scenarios, we conduct a significance test (t-test) for each dataset which is illustrated in Fig.~\ref{t_test}. Here, a significance level of 0.05 is applied as~\cite{DICD,DTLC}, and if the p-value is less than 0.05, the differences of accuracy between GDCAN and DCAN are statistically significant. For a clearer explanation, the $-\log(p)$ of each task has shown in the figure. We can also observe that majority of the $-\log(p)$ of the performance comparison between GDCAN and DCAN are larger than $-\log(0.05)$, which means GDCAN is significantly superior to DCAN in almost all scenarios.

As the difficulty of transfer tasks varies across different domains, we expect to study if the proposed adaptive routing strategy helps for harder tasks. Specifically, the t-test results on easy benchmarks ImageCLEF (4 out of 6) and Office-31 (4 out of 5) get slight significance while the results on hard benchmarks Office-Home (9 out of 12) and DomainNet (26 out of 30) obtain huge significance when comparing GDCAN and DCAN. This evidence validates that GDCAN is superior to DCAN when encountering large-scale challenging datasets and hard transfer tasks.

\vspace{-2mm}
\subsection{Generalized Results of DCAN and GDCAN}\label{sec:increment_exp}
To further demonstrate the generalization of our methods, we first adopt DCAN and GDCAN as incremental modules to two typical DA methods MSTN~\cite{MSTN} and CDAN~\cite{CDAN} and one domain-specialized DA method TransNorm~\cite{transnorm} without modifying their loss functions. Table~\ref{tab:increment1} and Table~\ref{tab:increment2} present the overall results on Office-Home and Office-31 datasets.
Specifically, in Office-Home, DCAN and GDCAN can generate $\mathbf{6.6\%}$ and $\mathbf{7.6\%}$ increases over MSTN, $\mathbf{5.8\%}$ and $\mathbf{6.5\%}$ increases over CDAN, and $\mathbf{4.5\%}$ and $\mathbf{6.3\%}$ increases over TransNorm. In Office-31, we also get comparable accuracies of DCAN and GDCAN, which improve the base DA methods by up to around $\mathbf{1.6\%}$. These gains imply that the designed structures in both DCAN and GDCAN can contribute to better adaptation performance when combining with the original methods.

Furthermore, we integrate DCAN and GDCAN into different CNN architectures i.e., ResNet-50 \cite{resnet} and ResNext-50 \cite{ResNext}, and train the combined CNN + DCAN/GDCAN with our proposed loss functions.
We can find that both DCAN and GDCAN can significantly enhance the capability of source-only networks. For instance, GDCAN can significantly surpass ResNet-50 by $\mathbf{13.6\%}$ on Office-31 and $\mathbf{25.4\%}$ on Office-Home in terms of average classification accuracy, and the improvement on ResNext-50 is from $\mathbf{8.9\%}$ to $\mathbf{14.7\%}$. By incorporating our methods into various CNN architectures, we can enable them to successfully mitigate the domain discrepancy and suitable for DA scenarios.

\begin{table*}[htbp] \footnotesize
  \centering
  \caption{\textbf{Ablation Study} of GDCAN on Office-Home for unsupervised DA.}\vspace{-4mm}
  \setlength{\tabcolsep}{0.4mm}{
    \begin{tabular}{lccccccccccccc}
    \toprule
    Method  & Ar $\rightarrow$ Cl  & Ar $\rightarrow$ Pr  & Ar $\rightarrow$ Rw  & Cl $\rightarrow$ Ar  & Cl $\rightarrow$ Pr  & Cl $\rightarrow$ Rw  & Pr $\rightarrow$ Ar  & Pr $\rightarrow$ Cl  & Pr $\rightarrow$ Rw  & Rw $\rightarrow$ Ar  & Rw $\rightarrow$ Cl  & Rw $\rightarrow$ Pr  & Avg. \\
    \hline
    GDCAN & \textbf{57.3}  & \textbf{75.7}  & \textbf{83.1}  & \textbf{68.6}  & 73.2  & \textbf{77.3}  & \textbf{66.7}  & \textbf{56.4}  & \textbf{82.2}  & \textbf{74.1}  & \textbf{60.7}  & \textbf{83.0}  & \textbf{71.5}  \\
    w/o $\mathcal{L}_\mathcal{M}^1$ + $\mathcal{L}_{reg}^1$ & 52.1  & 72.4  & 79.7  & 62.3  & 68.5  & 73.7  & 58.2  & 48.8  & 79.1  & 68.9  & 55.5  & 81.2  & 66.7  \\
    w/o $\mathcal{L}_\mathcal{M}^2$ + $\mathcal{L}_{reg}^2$ & 54.8  & 73.6  & 80.2  & 65.4  & 70.3  & 74.5  & 60.3  & 52.9  & 78.3  & 69.6  & 57.6  & 81.5  & 68.3  \\
    w/o $\mathcal{L}_{\mathcal{M}}$ + $\mathcal{L}_{reg}$ & 46.1 & 66.9 & 76.2 & 44.6 & 59.6 & 62.3 & 48.8 & 41.5 & 74.4 & 64.5 & 45.1 & 77.6 & 59.0   \\
    w/o  $\mathcal{L}_{reg}^1$ & 54.6  & 73.5  & 79.3  & 66.1  & 71.8  & 74.7  & 60.9  & 54.5  & 78.4  & 70.9  & 59.2  & 81.6  & 68.8  \\
    w/o  $\mathcal{L}_{reg}^2$ & 56.2  & 74.5  & 81.2  & 67.0  & 73.1  & 75.5  & 66.3  & 55.8  & 79.3  & 73.1  & \textbf{60.7}  & 82.2  & 70.4  \\
    w/o  $\mathcal{L}_{e}$ & 57.1  & 74.7  & 81.4  & 66.3  & 73.2  & 75.7  & 66.3  & 56.0  & 82.0  & 73.3  & 60.1  & 82.5  & 70.7  \\
    w/o  AAM & 54.2  & 74.1  & 79.5  & 64.9  & \textbf{74.4}  & 76.1  & 64.1  & 51.2  & 79.8  & 71.5  & 57.5  & 82.6  & 69.2    \\
    \bottomrule
    \end{tabular}%
    }
    \vspace{-4mm}
  \label{tab:ablation_home}%
\end{table*}%

\vspace{-2mm}
\section{Analysis}\label{sec:analysis}
\subsection{Ablation Study}\label{sec:ablation_study}
In order to examine the key components of our models, we perform ablation studies by removing each component from the whole framework at a time. Here, we take GDCAN as an example and mainly consider seven variants of GDCAN. Note that, for ResNet-50, we have two feature adaption modules (i.e., $L$ = 2): one after the average pooling layer, and the other after the softmax layer.
Therefore, we can obtain three variants: (1) ``GDCAN w/o $\mathcal{L}_\mathcal{M}^1+\mathcal{L}_{reg}^1$'' and (2) ``GDCAN w/o $ \mathcal{L}_\mathcal{M}^2+\mathcal{L}_{reg}^2$'', which denote respectively removing the corresponding adaptation module, and then remove all feature adaptation modules denotes as (3) ``GDCAN w/o $\mathcal{L}_\mathcal{M}$ + $\mathcal{L}_{reg}$''.
Moreover, the variants (4) ``GDCAN w/o $\mathcal{L}_{reg}^1$'' and (5) ``GDCAN w/o $\mathcal{L}_{reg}^2$'' mean the exclusion of regularization loss at different adaptation modules.
We additionally denote GDCAN without target entropy loss as (6) ``GDCAN w/o $\mathcal{L}_e$''.
Finally, to explore the effects of the adaptive attention module, we adapt the elimination of it, which is denoted as (7) ``GDCAN w/o AAM''.

The results of all GDCAN variants on Office-Home are illustrated in Table~\ref{tab:ablation_home}. It is clear that complete GDCAN outperforms other variants and gains large improvements.
Firstly, we can see that dramatic decreases happen in the average classification accuracy of ``GDCAN w/o $\mathcal{L}_\mathcal{M}^1+\mathcal{L}_{reg}^1$'' and ``GDCAN w/o $\mathcal{L}_\mathcal{M}^2+\mathcal{L}_{reg}^2$'', in which the former exhibits worse performance. This result reveals the effect of feature adaptation module is more important when getting closer to the lower layers. The same operation applies to the regularization loss, which can be affirmed by the fact that the accuracy of ``GDCAN w/o $\mathcal{L}_{reg}^1$'' is much lower than that of ``GDCAN w/o $\mathcal{L}^2_{reg}$''. Moreover, in DA, if there does not exist any alignment module (``GDCAN w/o $\mathcal{L}_\mathcal{M}$ + $\mathcal{L}_{reg}$ ''), it is hard to obtain expected performance on target data.

Secondly, it can also be observed that ``GDCAN w/o $\mathcal{L}_{e}$" achieves a relatively high accuracy to GDCAN, indicating entropy loss in target prediction can play a role as a complementary constraint. Moreover, it reveals that the adaptation performance of ``GDCAN w/o AAM'' suffers degradation of $\mathbf{2.3\%}$ w.r.t. average accuracy, manifesting the importance of the proposed adaptive attention module to explore the critical low-level domain-specialized knowledge.

In a nutshell, by jointly exploring the low-level domain conditioned channel attention and conducting high-level MMD-based feature alignment, the full GDCAN model can further benefit domain adaptation by reducing the cross-domain discrepancy underlying both the domain-specialized and task-specific distribution.

\begin{figure}[tb]
\centering
\includegraphics[width=0.85\columnwidth]{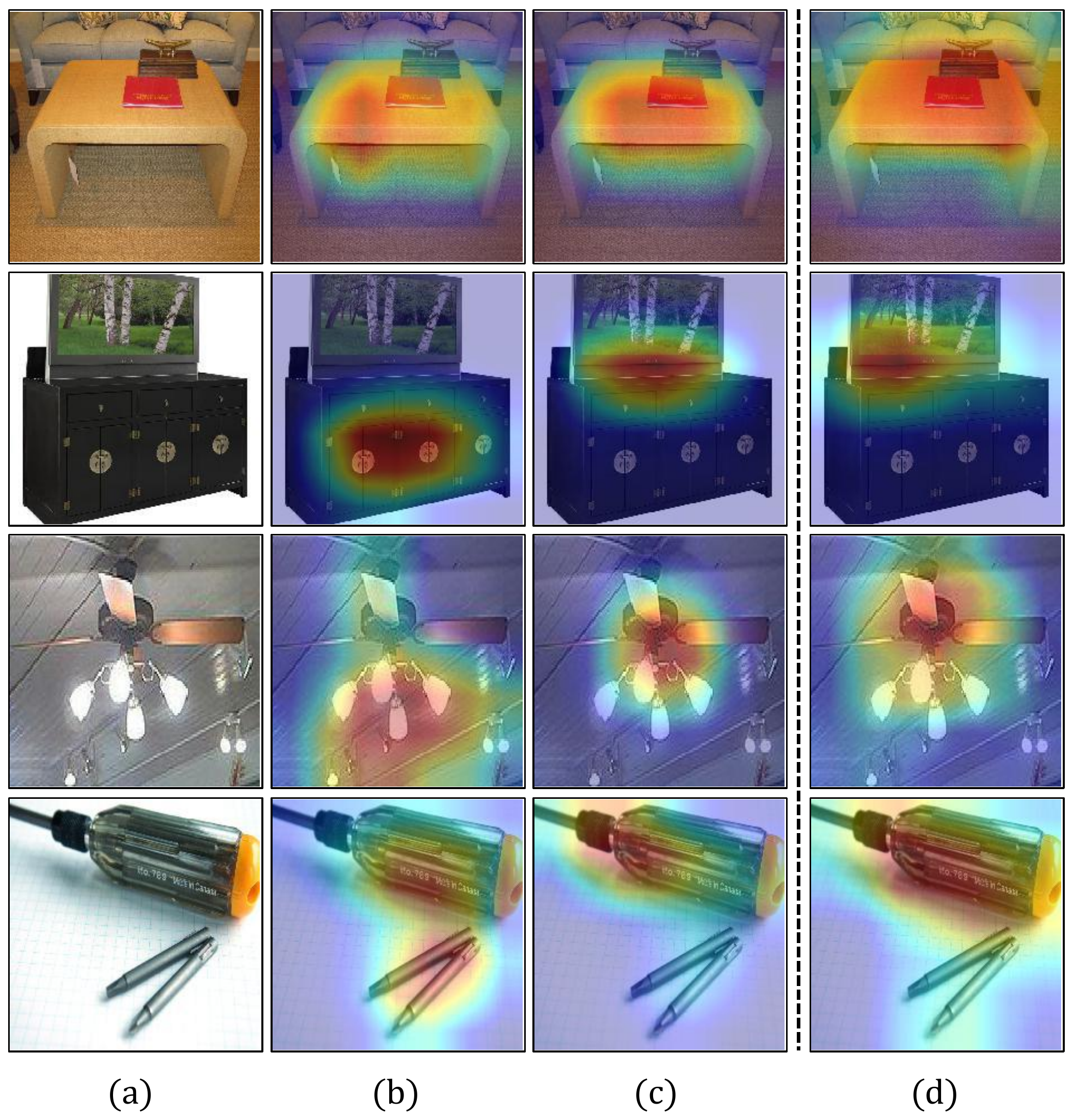}\vspace{-4mm}
\caption{Attention visualizations of the last convolutional layer learned by (b) DCAN, (c) GDCAN, and (d) target ground-truth model. The (a) column shows the original target images. (Best viewed in color.)}
\vspace{-4mm}
\label{Fig_attention_visual}
\end{figure}

\vspace{-2mm}
\subsection{Attention Visualization}
To explicitly validate the effectiveness of GDCAN in capturing domain-specific knowledge when compared with DCAN, as shown in Fig.~\ref{Fig_attention_visual}, we randomly select four target images and visualize attention maps of different models.

Given a target image in column (a), such as table, TV, fan, and screwdriver, ground-truth model can generate its corresponding attention map in column (d).
We find that DCAN sometimes cannot precisely locate the target region.
For example, in the $1^{st}$ row, although DCAN makes correct predictions, its attention map only responds to small local areas, whereas the attention maps of GDCAN and target ground-truth model are almost the same.
Moreover, from the $2^{nd}$ to $4^{th}$ rows, it is clear that DCAN concentrates on wrong objects.
Since GDCAN consistently highlights the most discriminative region that is desirable for target classification, the proposed adaptive attention module in GDCAN can capture domain-wise representations more effectively and achieve better performance.

\begin{figure}[tb]
  \centering
  \subfigure[Attention Value Difference]{
      \includegraphics[width=0.85\columnwidth]{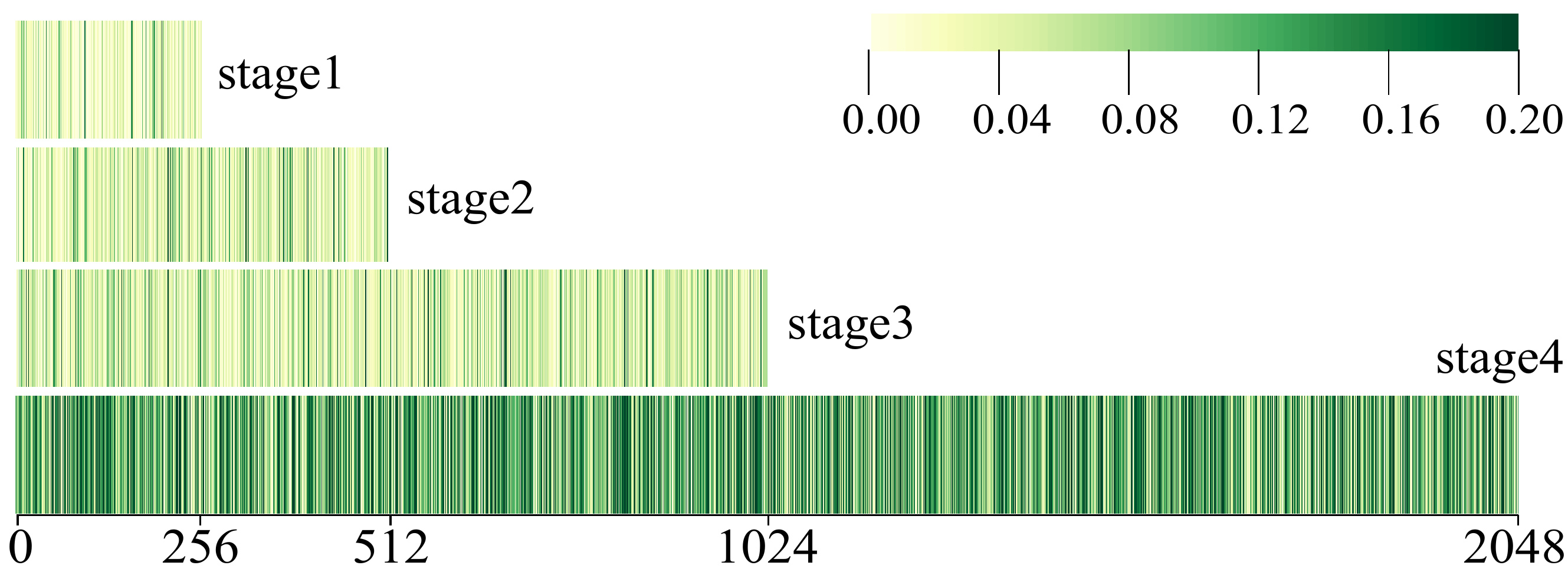}
      \label{Fig_attention1}
  }
  \subfigure[Attention Difference Comparison]{
      \includegraphics[width=0.85\columnwidth]{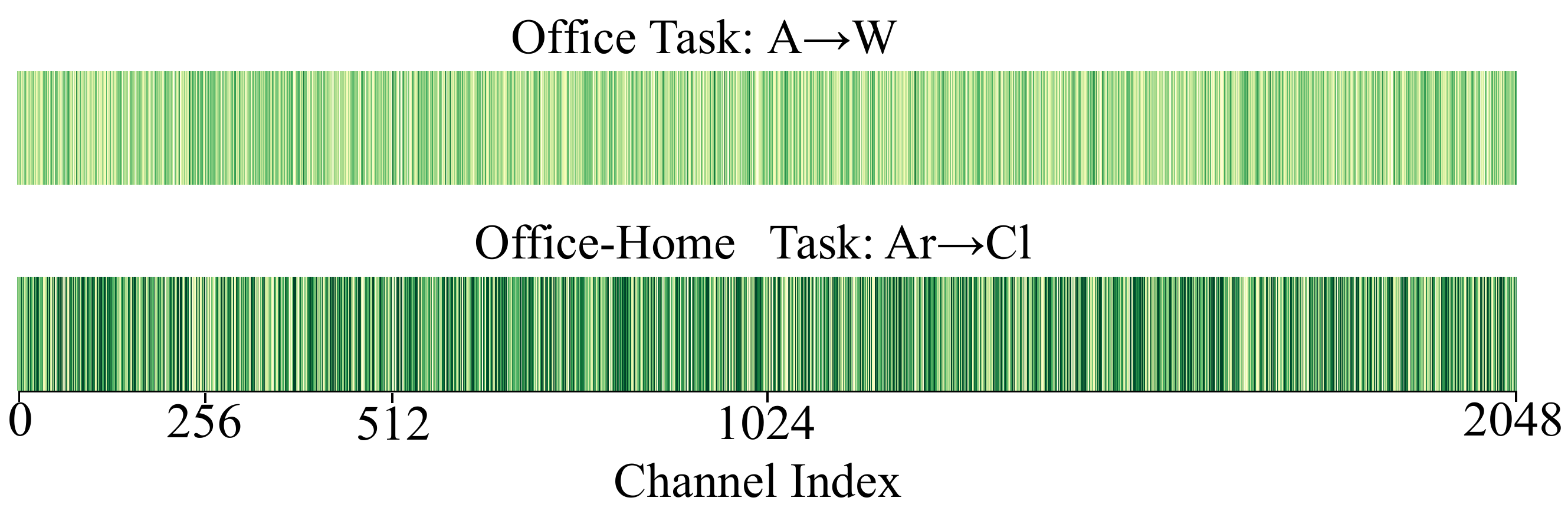}
      \label{Fig_attention2}
  }
  \vspace{-4mm}
  \caption{(a) The heat-map of attention value difference between source and target in our method on task Ar $\rightarrow$ Cl (Office-Home). The color of each vertical line represents the degree of attention difference across domains; (b) Attention difference comparison between task A $\rightarrow$ W (Office-31) and task Ar $\rightarrow$ Cl (Office-Home) at stage4.}
  \vspace{-4mm}
  \label{Fig_attention}
  \end{figure}

  \begin{table*}[htbp] \footnotesize
    \centering
    \caption{Analysis of adaptive routing strategy on Office-Home for unsupervised DA.}\vspace{-4mm}
    \setlength{\tabcolsep}{0.8mm}{
      \begin{tabular}{lccccccccccccc}
      \toprule
      Method  & Ar $\rightarrow$ Cl  & Ar $\rightarrow$ Pr  & Ar $\rightarrow$ Rw  & Cl $\rightarrow$ Ar  & Cl $\rightarrow$ Pr  & Cl $\rightarrow$ Rw  & Pr $\rightarrow$ Ar  & Pr $\rightarrow$ Cl  & Pr $\rightarrow$ Rw  & Rw $\rightarrow$ Ar  & Rw $\rightarrow$ Cl  & Rw $\rightarrow$ Pr  & Avg. \\
      \hline
      GDCAN & 57.3  & 75.7  & 83.1  & 68.6  & 73.2  & 77.3  & 66.7  & 56.4  & 82.2  & 74.1  & 60.7  & 83.0  & 71.5   \\
      w/ MMD & \textbf{59.5} & \textbf{76.7} & \textbf{83.8} & \textbf{69.1} & \textbf{73.7} & \textbf{77.8} & \textbf{68.1} & 57.1 & \textbf{82.8} & \textbf{74.5} & \textbf{62.0} & \textbf{83.7} & \textbf{72.4}  \\
      w/ KLD & 58.1 & 76.0 & 83.3 & 68.4 & 73.1 & 77.6 & 67.0 & \textbf{57.4} & 82.7 & 74.2 & 61.3 & 83.2 &  71.8  \\
      \hline
      GDCAN ($\lambda$ = 0.2) & 57.3  & 75.7  & 83.1  & 68.6  & 73.2  & 77.3  & 66.7  & 56.4  & 82.2  & 74.1  & 60.7  & 83.0  & 71.5   \\
      w/ adapt. $\lambda$ $\uparrow$ & 57.0 & 74.5 & 81.6 & 66.7 & 71.8 & 75.6 & 65.4 & 55.1 & 80.9 & 72.7 & 59.4 & 81.4 & 70.2   \\
      w/ adapt. $\lambda$ $\downarrow$ & \textbf{59.0} & \textbf{77.2} & \textbf{84.5} & \textbf{69.3} & \textbf{74.0} & \textbf{78.4} & \textbf{67.9} & \textbf{58.4} & \textbf{83.5} & \textbf{75.3} & \textbf{62.8} & \textbf{84.4} & \textbf{72.9}  \\
      \bottomrule
      \end{tabular}%
      }\vspace{-2mm}
    \label{tab:variant_home}%
  \end{table*}%

\vspace{-2mm}
\subsection{Channel Attention Difference Comparison}
As discussed in Section \ref{section:domain_specific_learning}, a major contribution in our method is proposing a domain adaptive channel attention module in convnets for better adaptability.
This structure aims to suppress noise while keeping useful information, and most importantly, excite specific channel values for each domain with multi-path design.
To provide a clear picture of this behavior, we present an intuitive way to visualize the channel attention values.

Given a backbone network ResNet-50, it consists of stages $m=\{1\,,2\,,3\,,4\}$ with regard to the numbers of channels in $\{256\,,512\,,1024\,,2048\}$. To understand the ability of the adaptive attention module, we calculate mean attention values of source and target samples in the last residual block of each stage, which are denoted as $\boldsymbol{\omega_s^{(m)}}$ and $\boldsymbol{\omega_t^{(m)}}$. For the $i$-th channel in stage $m$, we generate $\omega_i^{(m)}=|\omega_{si}^{(m)}-\omega_{ti}^{(m)}|$ to represent attention difference between domains. As shown in Fig. \ref{Fig_attention}, color brightness denotes difference magnitude. In other words, the brighter the color, the closer the channel activation value of source and target is under this channel.

\emph{1) Attention Value Difference:}
Fig. \ref{Fig_attention1} shows an example of attention difference in the last residual block across all stages.
Intuitively, the difference value becomes larger as the stage increases and the color in stage 4 is the darkest compared with other stages.
This performance is very similar to our statement that general representations lie in lower layers while domain-discriminative features are obtained in higher layers. It provides us new insights to design a more powerful deep convolutional structure for DA.
Therefore, instead of using the same convnets for both domains, our partially-shared structure would learn domain-wise channel response and improve cross-domain performance consistently.

\emph{2) Attention Difference Comparison:}
This experiment shows statistic comparison of attention module under tasks \textbf{A} $\rightarrow$ \textbf{W} (Office-31) and \textbf{Ar} $\rightarrow$ \textbf{Cl} (Office-Home) at stage4. In Fig. \ref{Fig_attention2}, it is easy to notice the channel activation difference from the bottom layer is much greater than that of the upper one. This means ``easy'' task collects global information of both domains, while ``hard'' task needs to model more specific channel attentions for each domain. In the meantime, it verifies our argument that capturing domain-discriminative features in convolutional layers is essential as well.

These results show that our model enjoys high efficiency and superior performance by making each domain have its own branch to learn specialized response for its features.

  \begin{figure}[tb]
  \centering
  \includegraphics[width=0.95\columnwidth]{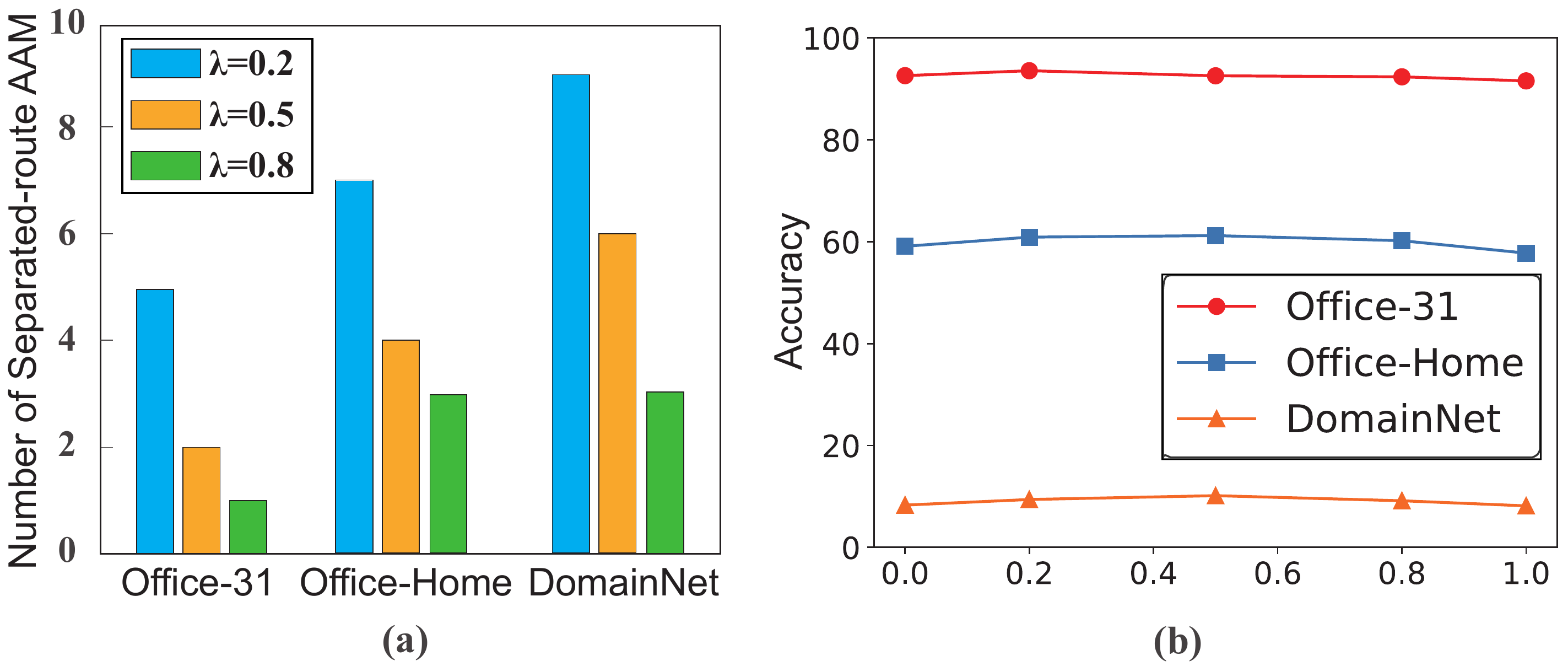}
  \vspace{-4mm}
  \caption{(a) Average number of route separation made by AAM, where the y-axis shows the number of AAMs that decide multi-path processing; (b) Parameter sensitivity analysis of $\lambda$ on tasks A $\rightarrow$ D (Office-31), Rw $\rightarrow$ Cl (Office-Home) and pnt $\rightarrow$ qdr (DomainNet).}
  \vspace{-4mm}
  \label{Fig_policies_visual}
\end{figure}

\begin{figure*}[tb]
  \centering
  \includegraphics[width=0.96\textwidth]{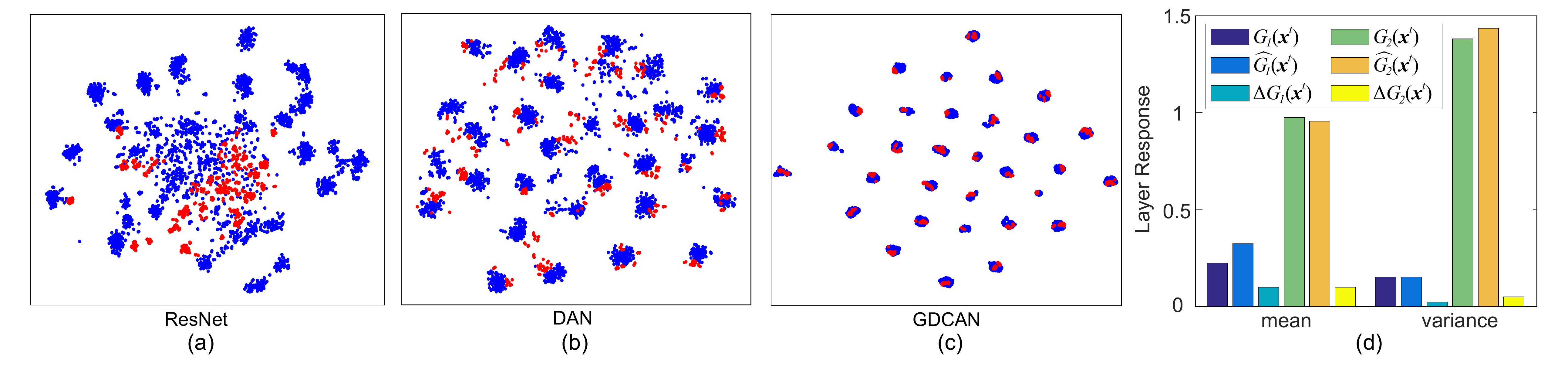}
  \vspace{-4mm}
  \caption{The t-SNE visualizations of (a) ResNet, (b) DAN and (c) GDCAN on task A $\rightarrow$ W of Office-31, where blue points are source data and red points are target data; (d) The statistics of response $G_1(\x^t)$, $\Delta G_1(\x^t)$, $\widehat{G_1}(\x^t)$, $G_2(\x^t)$, $\Delta G_2(\x^t)$ and $\widehat{G_2}(\x^t)$ on task A $\rightarrow$ W of Office-31.}
  \vspace{-4mm}
  \label{Fig_tsne}
  \end{figure*}

\vspace{-2mm}
\subsection{Adaptive Attention Module Analysis in GDCAN}~\label{adaptive_measurement}
\emph{1) Adaptive Attention Module}:
To better understand the adaptive routing strategy made by the adaptive attention module (AAM) in GDCAN, we randomly select three transfer tasks and count the average number of AAMs using route separation.
Note that since we use ResNet-50 as the backbone, a total of 16 AAMs are inserted.
Each AAM decides to apply route separation or just a single source route.

In Fig. \ref{Fig_policies_visual}(a), we find that at any threshold $\lambda$, the average number of AAMs using route separation increases with the difficulty rising of the cross-domain task.
In particular, the separation strategy is adopted the most frequently in DomainNet, while it is reversed in Office-31.
This phenomenon is reasonable, as route separation facilitates to capture more domain-specific features on hard tasks when the effect of general feature learning is limited. Besides, different $\lambda$ thresholds in $\{0.2\,,0.5\,,0.8\}$ are used for evaluation. Note that $\lambda=0$ denotes all separated-route structures in GDCAN, i.e., DCAN, while $\lambda=1.0$ represents all single-route structures.
It clearly shows that the number of AAMs using separation strategy decreases as the threshold $\lambda$ increases.
If the similarity of domain statistics is lower than threshold $\lambda$, we believe the domain difference is small enough to share one processing route.
A larger threshold means that the upper bound of the single route strategy is increased, so less separated routes would be triggered.

Meanwhile, we also report the impact of $\lambda$ on the classification accuracy by the parameter sensitivity.
From the analysis in Fig. \ref{Fig_policies_visual}(a), we know that the larger the $\lambda$, the more AAMs using a single route.
And in Fig. \ref{Fig_policies_visual} (b), we can see that on each task, the accuracy of GDCAN almost shows a concave curve with the increase of $\lambda$.
Specifically, in Rw $\rightarrow$ Cl of Office-Home, when $\lambda$ are 0, 0.2, 0.5, 0.8 and 1.0, the accuracies are $59.1\%$, $60.9\%$, $61.2\%$, $60.2\%$, and  $57.8\%$ respectively.
Moreover, in pnt $\rightarrow$ qdr of DomainNet, compared with $\lambda=0.5$, the performances of variants $\lambda=0.2\,,0.8$ are reduced by $0.8\%$ and $1.0\%$.
These results validate our hypothesis that it is more accurate to apply a flexible routing strategy rather than all single-route or all separated-route structures.

\emph{2) Adaptive Routing Strategy}: To enable a more concrete understanding about the adaptive version of GDCAN, we design two case studies. One is about options of cross-domain statistic distance $\widehat{m}$ in Section~\ref{sec:adaptive_strategy}. Here, we replace the original distance metric with another two widely used ones in DA, i.e., MMD (GDCAN w/ MMD) and kullback-leibler divergence (GDCAN w/ KLD). 

The other is about tactics for tuning the hyper-parameter $\lambda$. Since we use ResNet-50 as the backbone network, a total of 4 convolutional groups (conv2\_x, conv3\_x, conv4\_x, conv5\_x~\cite{resnet}) are included. we adopt two varied tactics, i.e., $\{0.2\,, 0.4\,, 0.6\,, 0.8\}$ (GDCAN w/ adapt. $\lambda$ $\uparrow$) or $\{0.8\,, 0.6\,, 0.4\,, 0.2\}$ (GDCAN w/ adapt. $\lambda$ $\downarrow$) for each AAM in different convolutional stage as its threshold, respectively.

Table~\ref{tab:variant_home} reports the classification accuracy results on Office-Home dataset. As can be seen from the $2^{nd}$ and $3^{rd}$ rows, the GDCAN model w/ MMD and w/ KLD slightly improve the average classification accuracy. The results demonstrate that our adaptive routing strategy is robust to the measurement of statistic distance across domains, and most of the commonly used distance measures are compatible. The results in the last two rows show that the threshold adjustment strategy with dynamic descent can get better performance, that is, knowledge transferability changes along convolutional layers~\cite{transferable}. As expected, the features are more general in the low-level layers, thus we allow source and target to share one single route with high probability. On the contrary, in the higher layers, the features are more task-specific thus we should enable triggering separated routes for source and target domains.

\vspace{-1mm}
\subsection{t-SNE Visualization}
As shown in Fig.~\ref{Fig_tsne}(a), (b), (c), we visualize t-SNE~\cite{tsne} embeddings of the features learned by ResNet-50, DAN, and GDCAN respectively. Note that each class is denoted as a cluster and different domains are in different colors.
It is clear that the features of ResNet cannot align the distributions well, especially the disorder distribution of target samples without forming obvious inter-class boundaries.
DAN is capable of obtaining more compact features than ResNet, whereas some samples still scattering around clusters.
Compared with them, GDCAN shows a better ability to make inter-class separated and intra-class clustered tightly, which reveals the proposed components can promote the network to learn highly discriminable representations.

\vspace{-1mm}
\subsection{Layer Response}
We illustrate the efficacy of the feature adaptation modules by computing the mean and variance of two task-specific layer outputs in this experiment.
If the plugged modules learn some domain deviation information, there should be some layer responses reflected by the mean and variance values.
As shown in Fig. \ref{Fig_tsne}(d), since there exist layer activations in adaptation modules, the designed structure can respond to the inputs.
Moreover, we observe that $\Delta G_1(\x^t)$ and $\Delta G_2(\x^t)$ could automatically learn the domain discrepancies instead of zero-response, which suggests the adaptation modules can facilitate the precise feature correction and discriminative knowledge transfer.

\begin{figure}[tb]
  \centering
  \includegraphics[width=0.95\columnwidth]{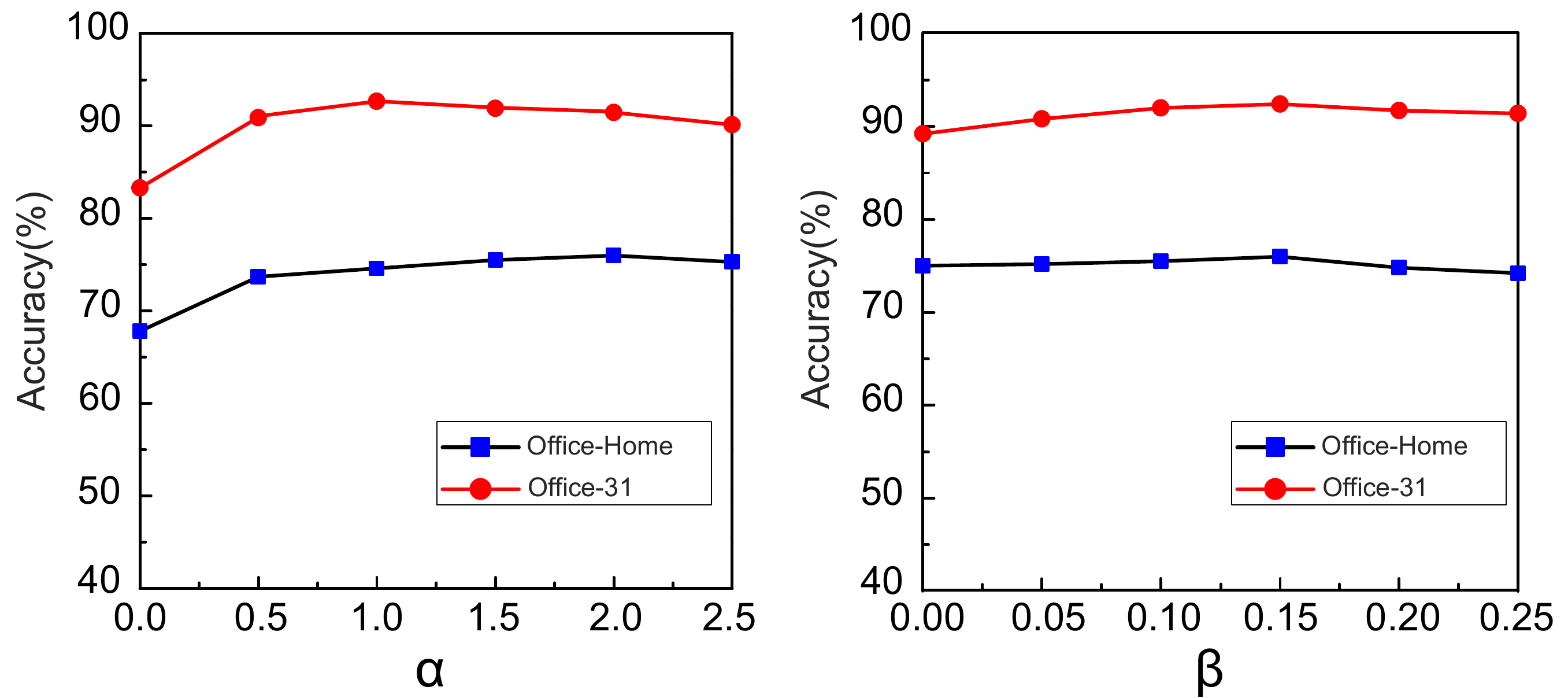}
  \vspace{-4mm}
  \caption{Parameter sensitivity analysis of $\alpha$ and $\beta$ on task Ar $\rightarrow$ Pr (Office-Home) and task A $\rightarrow$ D (Office-31).}
  \vspace{-4mm}
  \label{Fig_parameter}
\end{figure}

\vspace{-1mm}
\subsection{Parameter Sensitivity}
We conduct parameter sensitivity analysis to evaluate the sensitivity of GDCAN on tasks Ar $\rightarrow$ Pr (Office-Home) and A $\rightarrow$ D (Office-31). As shown in Fig. \ref{Fig_parameter}, we select balance weights from $\alpha \in \{0\,,0.5\,,1\,,1.5\,,2\,,2.5\}$ and $\beta \in \{0\,,0.05\,,0.1\,,0.15\,,0.2\,,0.25\}$. For $\alpha$, the accuracy of GDCAN increases first and then decreases slightly. Similarly, as $\beta$ gets larger, the performance presents a slow bell-shaped curve as well. In particular, we observe that the lowest accuracy when $\alpha=0$ or $\beta = 0$. This confirms the validity of alignment and entropy penalties in Eq.~\eqref{objective}. Therefore, it is necessary to set proper weight for each penalty. In addition, the overall performance of our method would not be greatly influenced by the value of trade-off parameters, which indicates GDCAN is not quite sensitive to $\alpha$ and $\beta$.

\vspace{-1mm}
\section{Conclusion}\label{sec:conclusion}
In this paper, we presented a Generalized Domain Conditioned Adaptation Network (GDCAN) to simultaneously achieve domain-specialized feature learning in low-level convolutional features and effectively mitigate distribution mismatch by domain-invariant feature learning at higher levels.
Unlike the previous completely-shared convolutional scheme for DA, we replaced it with a partially-shared one that is implemented with the domain conditioned channel attention module.
This structure is equipped with an adaptive routing strategy to precisely capture domain-specific knowledge in low-level so as to facilitate subsequent feature migration. In the higher stage, we used the feature adaptation module guided by regularization at several task-specific layers for effective domain gap mitigation. Moreover, GDCAN can be used as an incremental module to significantly enhance the feature transferability of popular CNNs and other DA methods. We conducted extensive experiments on four datasets, and our GDCAN presented significant improvements over the state-of-the-art models, especially when it came to very tough cross-domain tasks.

\vspace{-2mm}
\section*{Acknowledgements}
This work is supported in part by the National Natural Science Foundation of China under Grant No. 61902028, and in part by the National Key Research and Development Plan of China under Grant No. 2018YFB1003701 and 2018YFB1003700.

\ifCLASSOPTIONcaptionsoff
  \newpage
\fi



%
\bibliography{reference}
\bibliographystyle{IEEEtran}

\vspace{-10mm}
\begin{IEEEbiography}[{\includegraphics[width=1in,height=1.25in,clip,keepaspectratio]{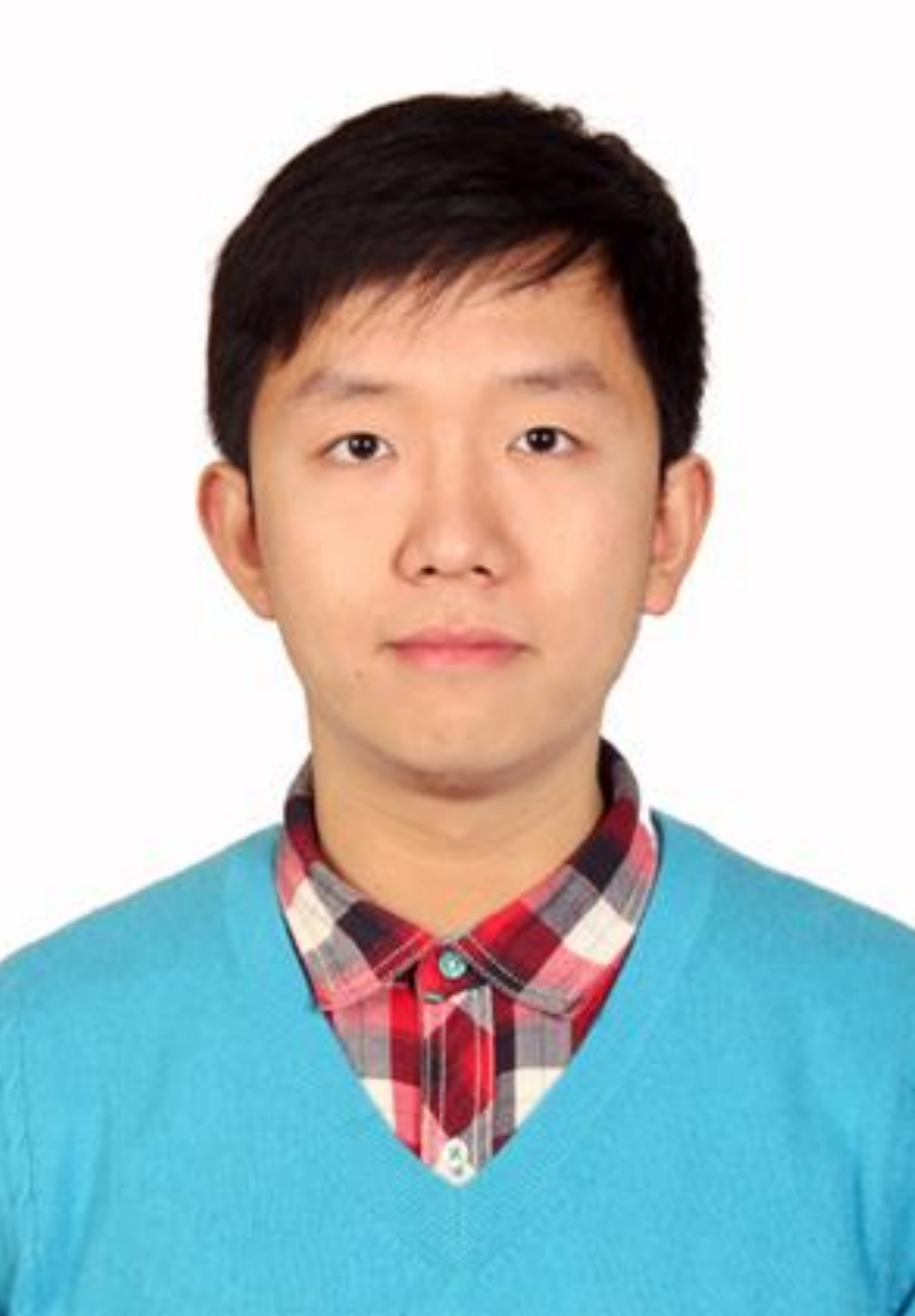}}]{Shuang Li} received the Ph.D. degree in control science and engineering from the Department of Automation, Tsinghua University, Beijing, China, in 2018.

He was a Visiting Research Scholar with the Department of Computer Science, Cornell University, Ithaca, NY, USA, from November 2015 to June 2016. He is currently an Assistant Professor with the school of Computer Science and Technology, Beijing Institute of Technology, Beijing. His main research interests include machine learning and deep learning, especially in transfer learning and domain adaptation.
\end{IEEEbiography}
\vspace{-10mm}

\begin{IEEEbiography}[{\includegraphics[width=1in,height=1.25in,clip,keepaspectratio]{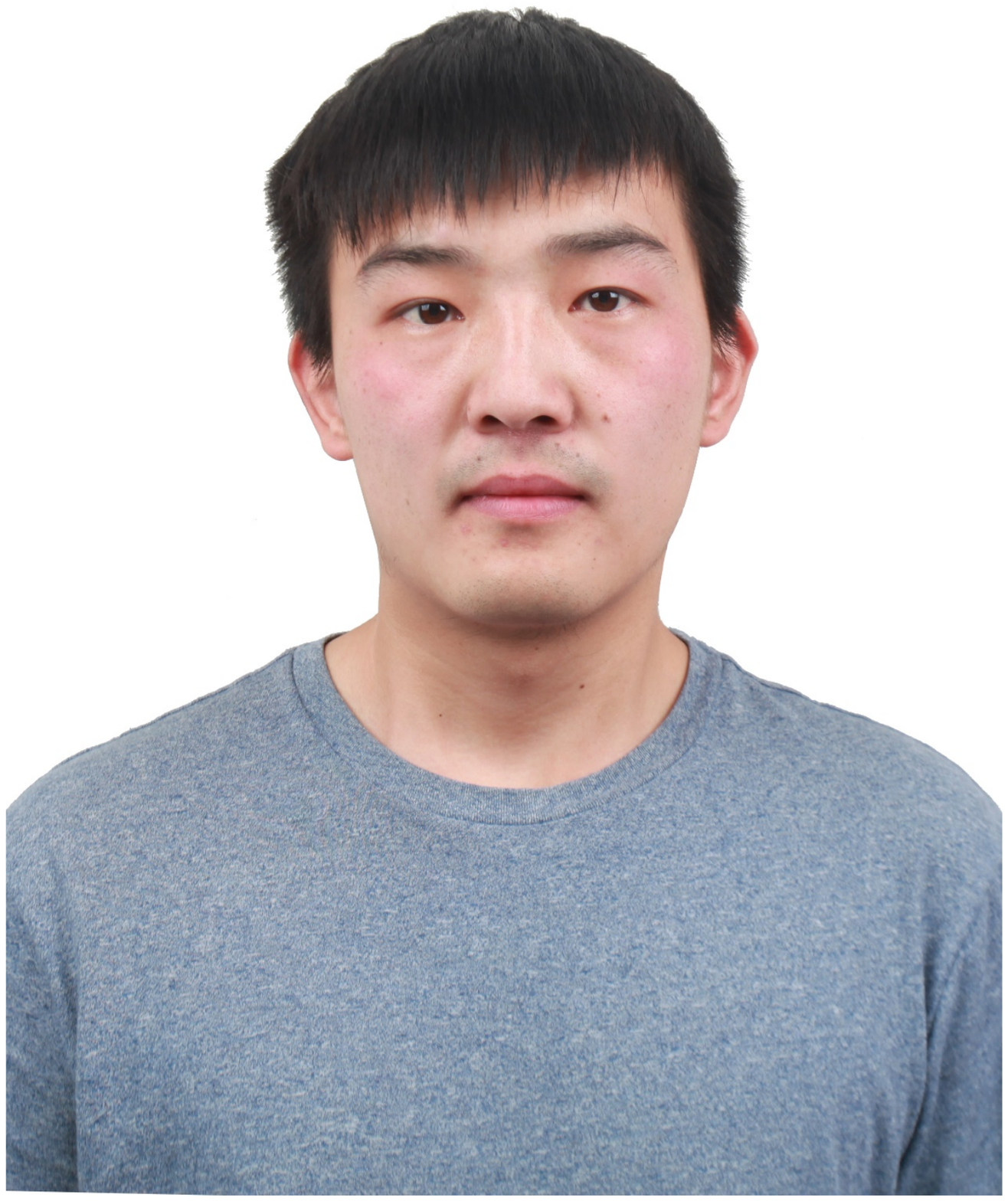}}]{Binhui Xie} is a graduate student at the School of Computer Science and Technology, Beijing Institution of Technology. His research interests focus on computer vision and transfer learning.
\end{IEEEbiography}
\vspace{-10mm}

\begin{IEEEbiography}[{\includegraphics[width=1in,height=1.25in,clip,keepaspectratio]{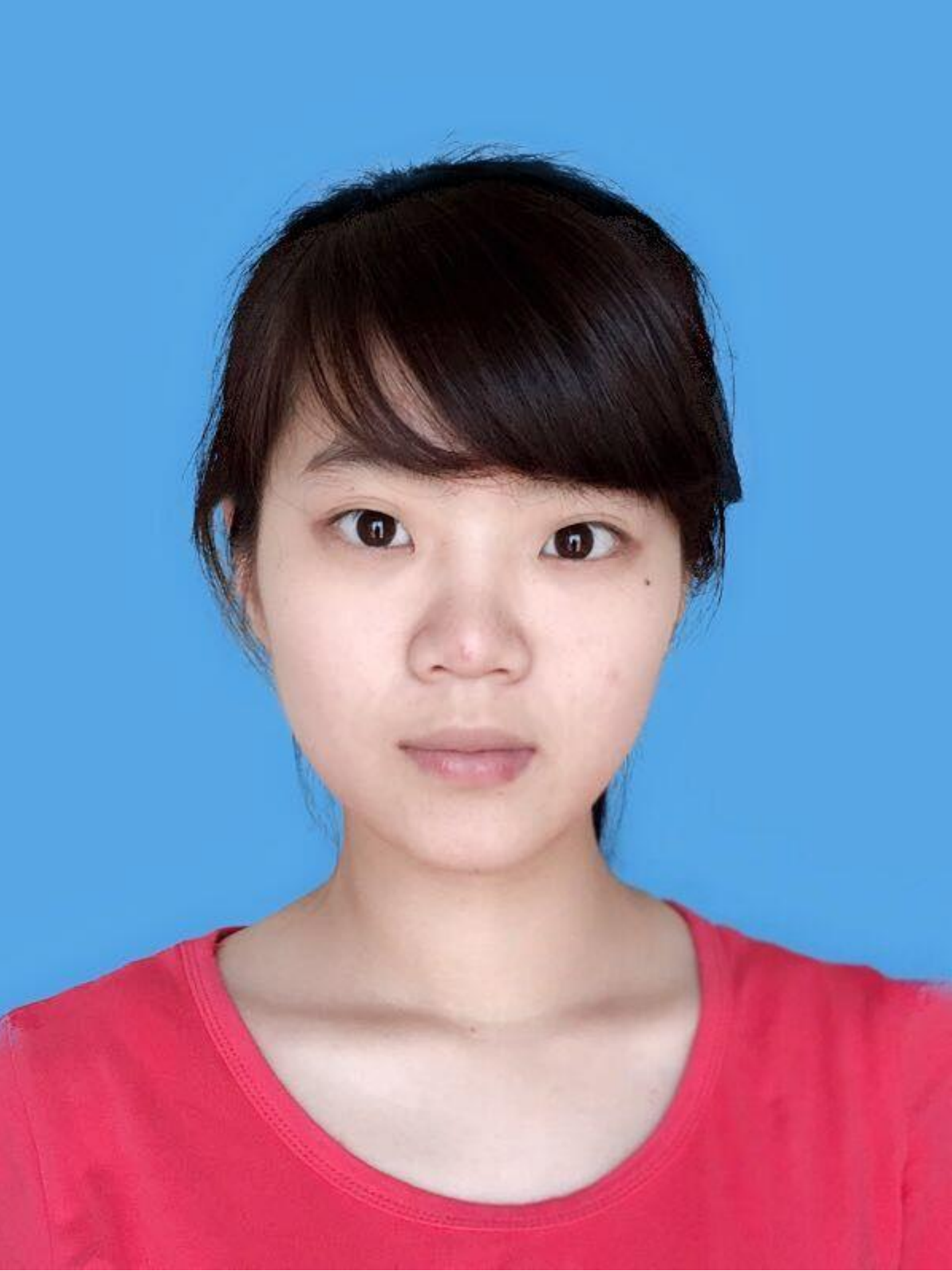}}]{Qiuxia Lin} is pursuing the M.S. degree in Computer Science from Beijing Institute of Technology. Her research interests include deep learning and transfer learning.
\end{IEEEbiography}
\vspace{-10mm}

\begin{IEEEbiography}[{\includegraphics[width=1in,height=1.25in,clip,keepaspectratio]{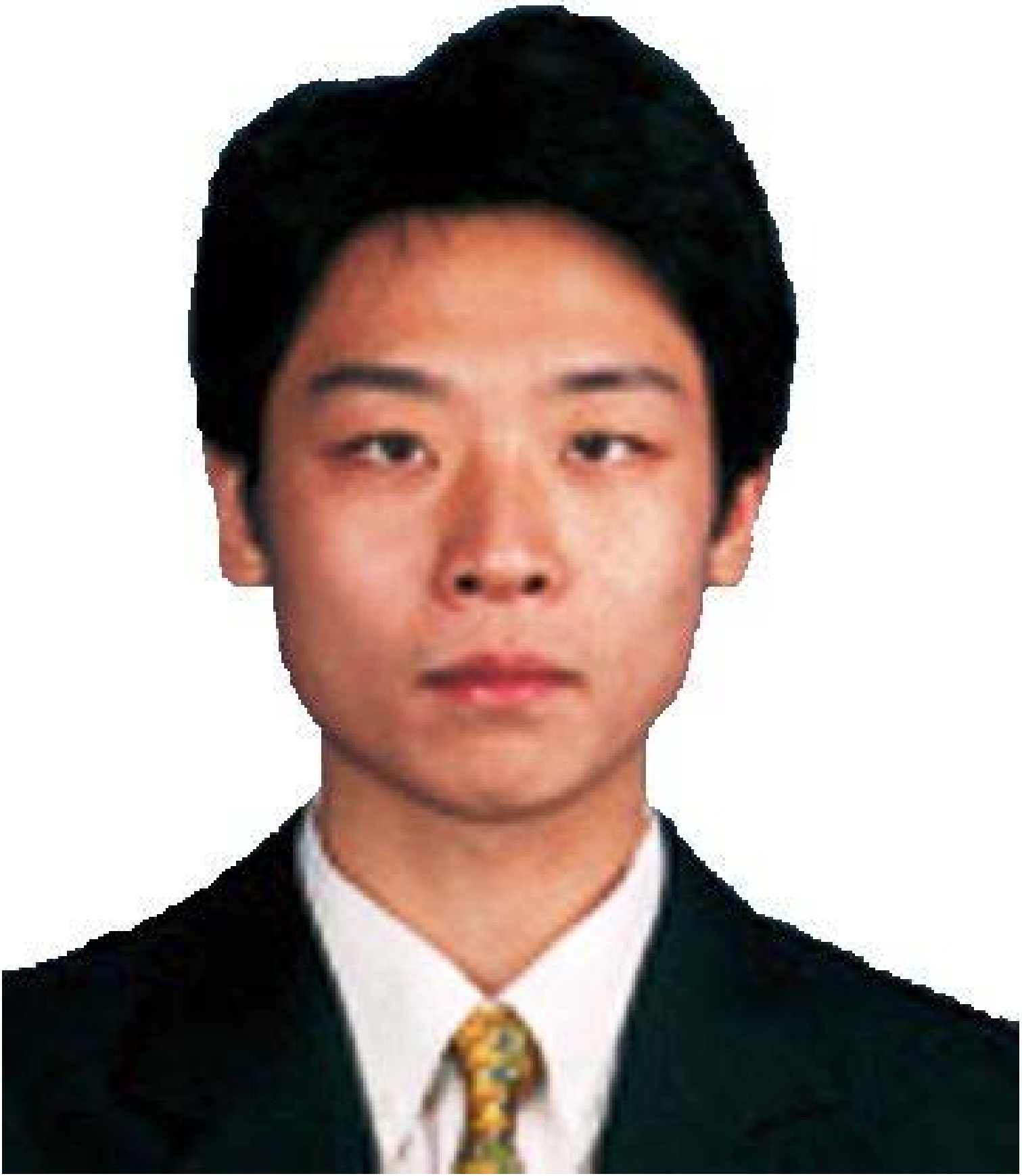}}]{Chi Harold Liu} receives the Ph.D. degree from Imperial College, UK in 2010, and the B.Eng. degree from Tsinghua University, China in 2006.

He is currently a Full Professor and Vice Dean at the School of Computer Science and Technology, Beijing Institute of Technology, China. Before moving to academia, he joined IBM Research - China as a staff researcher and project manager, after working as a postdoctoral researcher at Deutsche Telekom Laboratories, Germany, and a visiting scholar at IBM T. J. Watson Research Center, USA. His current research interests include the Big Data analytics, mobile computing, and deep learning. He has published more than 90 prestigious conference and journal papers and owned more than 14 EU/U.S./U.K./China patents. He is a Fellow of IET, and a Senior Member of IEEE.
\end{IEEEbiography}
\vspace{-10mm}

\begin{IEEEbiography}[{\includegraphics[width=1in,height=1.25in,clip,keepaspectratio]{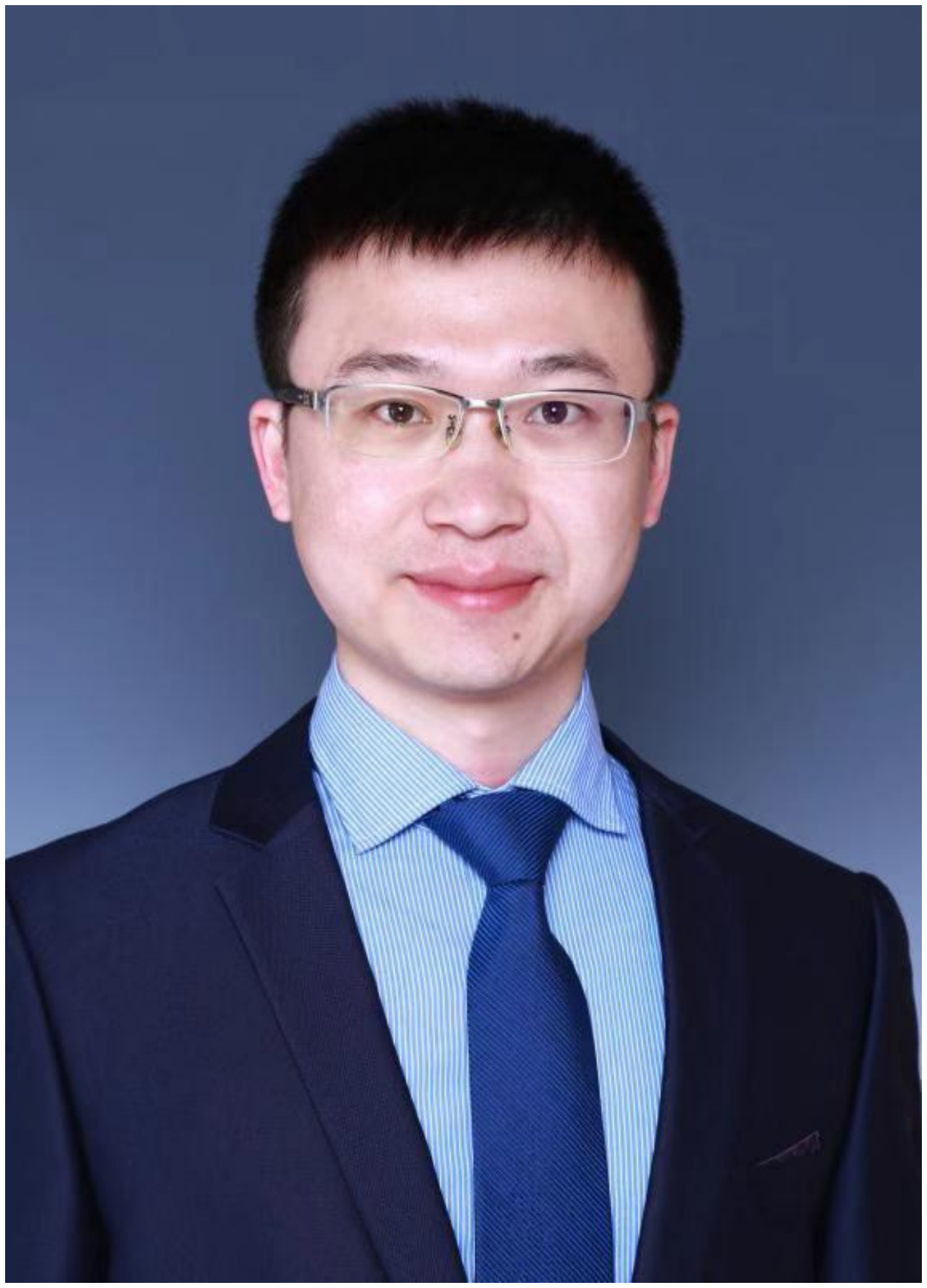}}]
{Gao Huang} is an assistant professor in the Department of Automation, Tsinghua University. He was a Postdoctoral Researcher in the Department of Computer Science at Cornell University. He received the PhD degree in Control Science and Engineering from Tsinghua University in 2015, and B.Eng degree in Automation from Beihang University in 2009. He was a visiting student at Washington University at St. Louis and Nanyang Technological University in 2013 and 2014, respectively. His research interests include machine learning and computer vision.
\end{IEEEbiography}
\vspace{-10mm}


\begin{IEEEbiography}[{\includegraphics[width=1in,height=1.3in,clip,keepaspectratio]{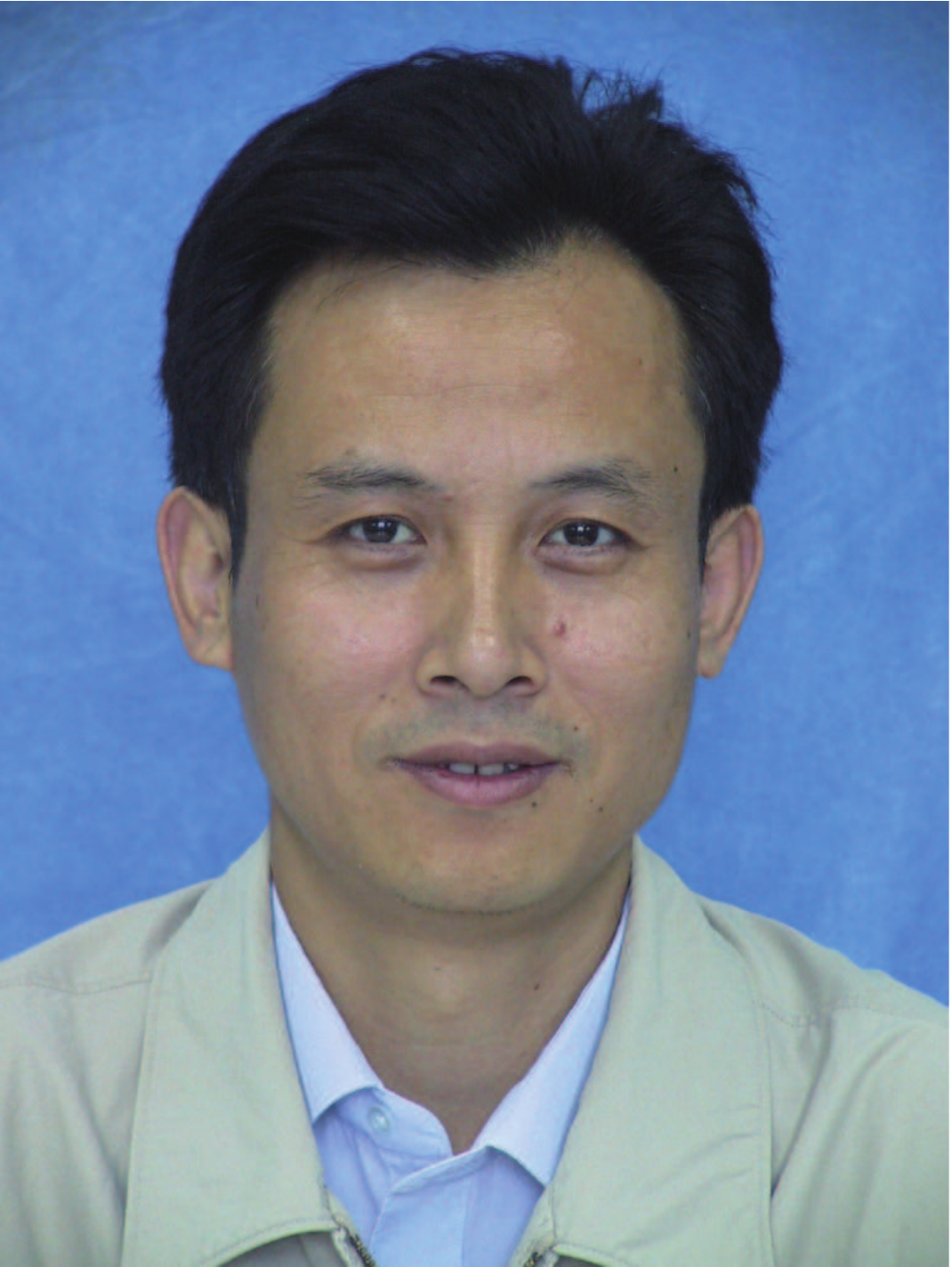}}]{Guoren Wang} received the BSc, MSc, and PhD degrees from the Department of Computer Science, Northeastern University, China, in 1988, 1991 and 1996, respectively. Currently, he is a Professor and the Dean with the School of Computer Science and Technology, Beijing Institute of Technology, Beijing, China. His research interests include XML data management, query processing and optimization, bioinformatics, high dimensional indexing, parallel database systems, and cloud data management. He has published more than 100 research papers.
\end{IEEEbiography}

\end{document}